\RequirePackage{fix-cm}
\documentclass[twocolumn,natbib]{svjour3}          
\smartqed  
\usepackage{graphicx}
\usepackage{enumerate}
\usepackage{graphics}
\usepackage{multirow}
\usepackage{array}
\usepackage{booktabs}
\usepackage{amsfonts}
\usepackage{amsmath}
\usepackage{amssymb}
\usepackage{esvect}

\usepackage{centernot} 

\usepackage{subfigure} 

\usepackage{algorithm}
\usepackage[noend]{algpseudocode}

\makeatletter
\def\BState{\State\hskip-\ALG@thistlm}
\makeatother

\newcommand{\ie}{\textit{i}.\textit{e}.}
\newcommand{\Ie}{\textit{I}.\textit{e}.}
\newcommand{\eg}{\textit{e}.\textit{g}.}

\journalname{}
\begin{document}

\title{Global Semantic Description of Objects based on Prototype Theory
}


\author{Omar  Vidal Pino         \and
        Erickson R. Nascimento   \and
        Mario F. M. Campos 
}


\institute{Omar Vidal Pino \and Erickson~R.~Nascimento \and
	Mario~F.~M.~Campos \at
              Universidade Federal de Minas Gerais, Computer Science Department, Computer Vision and Robotics Lab, Belo Horizonte, Minas Gerais, Brazil \\
              Tel.: +55-31-3409 5856 \\
              Fax:  +55-31-3409 5858 \\
              \email{\{ovidalp, erickson, mario\}@dcc.ufmg.br.}           
}

\date{Received: date / Accepted: date}

\maketitle
\begin{abstract}

In this paper, we introduce a novel semantic description approach based on Prototype Theory foundations. Inspired by the human approach used for representing categories, we propose a Computational Prototype Model~(CPM) that \textit{encodes} and \textit{stores} the central semantic meaning of the object's category: the semantic prototype. Also, we introduce a Prototype-based Description Model that encodes the semantic of an object while describing its features using our CPM model. Our description method uses semantic prototypes computed by convolutional neural network (CNN) classification models to create discriminative signatures that describe an object highlighting its most distinctive features within the category. Our experiments show that: \textit{i)} the proposed CPM model (semantic prototype + distance metric) successfully describes the internal semantic structure of objects categories; \textit{ii)} our semantic distance metric can be understood as object visual typicality score within a category; and \textit{iii)} our descriptor encoding is semantically interpretable and significantly outperforms other image global encodings in clustering and classification tasks.




\end{abstract}
%
%

\section{Introduction}
\label{sec:introduction}
Memory is one of the most amazing faculties of the human being. It is generally considered as the brain ability to \textit{code}, \textit{store}, and \textit{retrieve} information~\citep{atkinson1968}. \textit{Semantic memory}~\citep{tulving2007coding}, for instance, refers to general world knowledge that we accumulate throughout our lives~\citep{mcrae2013}. A relevant aspect of the functional neuroanatomy of the semantic memory resides in the \textit{representation of the meaning} of objects and their properties~\citep{martin2007representation}. Several assumptions indicate that  human beings are capable of: \textit{i)}~learning the most distinctive features of a specific object category~\citep{martin2007representation,thompson2003neuroimaging}; \textit{ ii)}~form categories and~\textit{object semantic definitions}~(abstractions) at a very early age~\citep{martin2007representation}. Semantic memory involves the semantic definition of objects~\citep{tulving2007coding} and, consequently, the success of object recognition, classification, and description tasks are causally related to the success of effectively recovering the learned knowledge~\citep{tulving2007coding}.

For several years, the fields of Computer Vision and Machine Learning have tried to build and learning pattern recognition methods with a similar performance of a human being for visual information processing. Although the state-of-the-art methods have achieved surprising results, there are still many challenges to achieve the discriminative power and abstraction of semantic memory to represent the semantic. How to describe and stand for objects, semantically? How to simulate the behavior of semantic memory in the representation of learned knowledge of objects' features? How to extract and encode the object features to encapsulate the representation of the meaning (or \textit{semantic representation}) of a specific object? How to learn the semantic definition of categories objects and use this definition in object recognition, classification, and description tasks? These are just some of the interesting questions that still occupy the investigation agenda of many research areas.

\begin{figure*}[t]
	\begin{center}
		
		\includegraphics[width=0.98\linewidth]{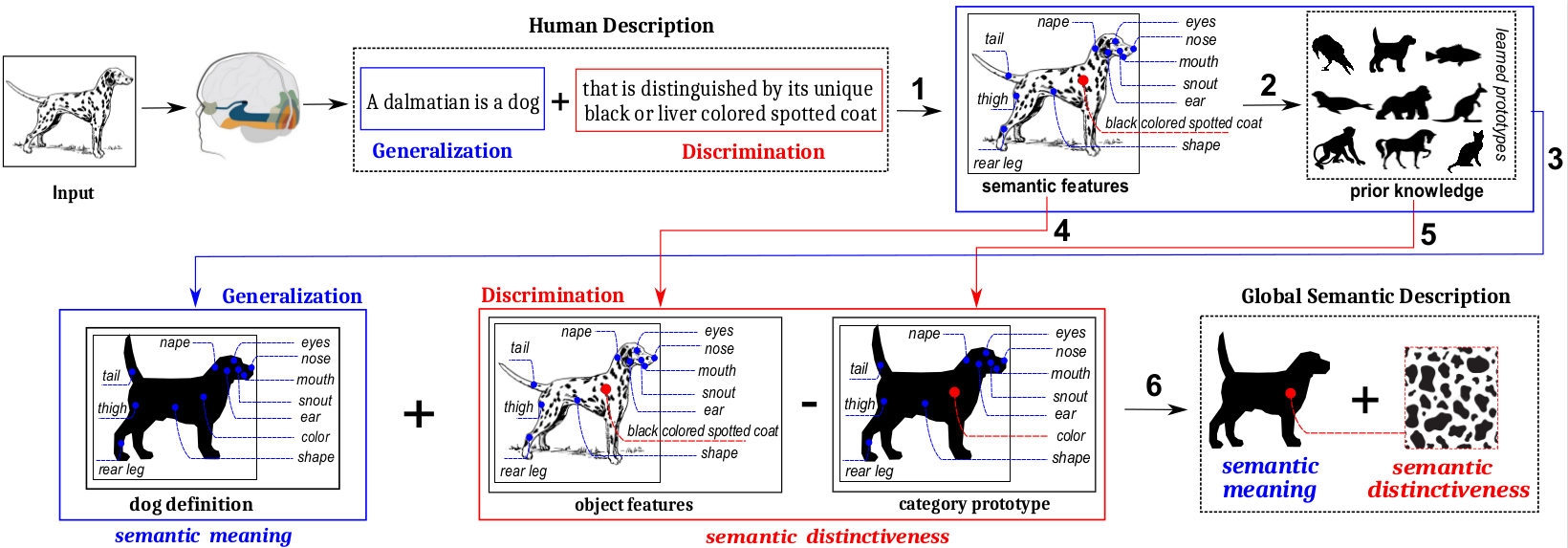}
	\end{center}
	\caption{\textit{Motivation and Concepts}. Schematic of our prototype-based description model. The human visual system can observe an object and to build an object semantic description that highlighting their most distinctive features within the object category. We propose a prototype-based model to simulate this behavior through the processing flow from  \textit{1)} to \textit{6)}.
		\textit{1)} features extraction; \textit{2)} object features recognition; \textit{3)} categorization; \textit{4)} object features; \textit{5)} central semantic meaning of a  category~(the category prototype); \textit{6)} our Global Semantic Description based on Prototypes.}
	\label{fig:motivation}
\end{figure*}

In this paper -- motivated by the semantic memory behavior -- we propose a mathematical model that attempts to represent the semantic definition of object categories. Also, we propose a procedure to introduce this semantic representation of object categories in the global description of objects features extracted from images.



The knowledge extraction models (high-level vision processes) from images are highly influenced by the methods used for detection, extraction, and representation of image relevant information. Consequently, the extraction of image relevant features has been the subject of Computer Vision research for decades. For several years, hand-crafted features~\citep{bay2008SURF,lowe2004SIFT,tola2008DAISY} and machine learning methods~\citep{simonyan2014learning,strecha2012ldahash} were the choice for image feature description tasks. 

The advent of Convolutional Neural Networks~(CNN) outperformed these traditional methods and enabled them to achieve a visual recognition model with similar behavior of \textit{semantic memory} for classification tasks \citep{he2016deep,simonyan2014very,szegedy2017inception}, sparking the tendency of images semantic processing with deep-learning techniques. The CNN-models success spawned numerous CNN-descriptors produced with different approaches that learn effective representations for describing image features \citep{han2015matchnet,kim2017fcss,simo2015discriminative,zagoruyko2015learning}. 
Consequently, representations of image features extracted using deep classification models \citep{he2016deep,simonyan2014very,szegedy2017inception}, or using  CNN-descriptors are commonly referred as \textit{semantic feature} or \textit{semantic signature}. 

\textit{Semantic feature} term has been extensively studied in the field of linguistic semantic; it is defined as the representation of the basic conceptual components of the meaning of any lexical item \citep{fromkin2018introduction}. In the seminal work of \cite{rosch1975cognitive}, the author analyzed the semantic structure of the meaning of words and introduced the concept of \textit{semantic prototype}.
~According to \cite{rosch1975cognitive,rosch1975family}, the representation of \textit{category semantic meaning} is related to the \textit{category prototype}, particularly to those categories naming natural objects.

Image semantic understanding is influenced by how are semantically represented the features of image basic components~(\eg, objects), and the semantic relations between these basic components~\citep{guo2016deep}. CNN-description models \citep{han2015matchnet,lin2016learning,simo2015discriminative,zagoruyko2015learning} and semantic description models~\citep{han2017scnet,kim2017fcss,Rocco18} stand for the semantic information of image features using a range of different approaches. Nevertheless, \textit{none} of these models codify the representation of the visual information based on the theoretical foundation of Cognitive Science to represent the \textit{semantic meaning}.


In this paper, we rely on cognitive semantic studies related to the Prototype Theory for modeling the \textit{central semantic meaning} of objects categories: the prototype.
We propose a novel approach to take on the semantic features descriptions of objects based on prototypes. Our \textit{prototype-based description model} uses the category's prototype to find a global semantic representation of the basics conceptual components (objects) of the image meaning.

To achieve this goal, we bring to light the Prototype Theory as a theoretical foundation to represent the semantic meaning of the visual information accurately to represent --- semantically ---  the basics components of the image: objects. The Prototype Theory proposes that human beings think a category in terms of abstract prototypes, defined by typical members of the category~\citep{geeraerts2010theories,rosch1975cognitive,rosch1978principles}. This theory also exposes that successful execution of object recognition and description tasks in the human brain is inherently related to the learned prototype of the object category~\citep{minda2002comparing,rosch1975cognitive,rosch1978principles,zaki2003prototype}.
The observations on the Prototype Theory raise the following two questions: i) Can a model of the perception system be developed in which objects are described using the same semantic features that are learned to identify and classify them? ii) How can the category prototype be included in the object global semantic description?
%



We address these two questions motivated by the human's approach to describing objects globally. Human being uses the generalization and discrimination processes to build object descriptions that highlighting their most distinctive features within the category. 
For example, a typical human description: a dalmatian is a dog~(generalization ability to recognize the central semantic meaning of dog category) that is distinguished by its unique black, or liver-colored spotted coat~(discrimination ability to detect the semantic distinctiveness of object within the dog category).~Figure~\ref{fig:motivation} illustrates the intuition and principal concepts of our prototype-based description model. The main idea of our approach is to use the quality of features extracted with CNN-classification models both to represent the central semantic meaning of a specific category and learn the object distinctiveness within the category.

More specifically, our main contributions in this paper are as follows:
\begin{enumerate}
	\item a \textit{Computational Prototype Model}~(CPM) based on Prototype Theory foundations, to stand for the central semantic meaning of object images categories.  Our CPM model allows to interpret  possible semantic associations between members  within the category internal structure.
	
	\item a \textit{semantic distance metric} in object image CNN features domain, which can be understood as a measure of object typicality within the object category. 
	\item a \textit{prototype-based description model} for global semantic description of objects images. Our semantic description model introduces, for the first time, the use of category prototypes in image global description tasks.
\end{enumerate}





\section{Related works}
\label{sec:related_work}

\subsection*{CNN descriptors}
Descriptors extracted using CNN techniques have shown that it is possible, for a learning approach, to outperform the best techniques based on carefully hand-crafted features~\citep{bay2008SURF,lowe2004SIFT,tola2008DAISY}. CNN descriptor models differ among themselves on how to compute the descriptors in their deep architectures, similarity functions learning, and its features extraction methods. Some approaches extract immediate activations of the model as a descriptor signature~\citep{simonyan2014very,szegedy2017inception, donahue2014decaf,long2014convnets}. Others methods use similarity convolutional networks~\citep{han2015matchnet,simo2015discriminative,yi2016lift,zagoruyko2015learning} and Siamese networks~\citep{han2015matchnet,zagoruyko2015learning,yi2016lift} to learn discriminative representations.~LIFT~\citep{yi2016lift} learns each task involved in features management: detection, orientation estimation, and description. \cite{lin2016learning} constructed a compact binary descriptor for efficient object matching based on features extracted with VGG16 model~\citep{simonyan2014very}. Those CNN-descriptor models were more oriented to achieve discriminative features than representing the image semantic. 
\subsection*{Semantic descriptors and semantic correspondence}
\cite{liu2011sift} proposed SIFT Flow method. SIFT Flow method generated the start of semantic flow family methods as a solution to the challenge of semantic correspondence~\citep{bristow2015dense,liu2011sift, yang2014daisy}. Several of these methods combine their approaches with the extraction of hand-crafted features~\citep{lowe2004SIFT,tola2008DAISY}. 
Some works~\citep{han2017scnet,kim2017fcss, Rocco18} use the robustness of CNN-models for training deep learning architectures and address the problem of semantic correspondence. \cite{kim2017fcss} tackled the problem of semantic correspondence by constructing FCSS semantic descriptor. 
In general, CNN descriptors and semantic descriptors are trained to learn their semantic representations and use different deep learning architectures. Most of these features description models do not use the discriminative power of the features extracted using the well-known CNN-classification models~\citep{he2016deep,simonyan2014very,szegedy2017inception}. Moreover, none of these CNN-feature description approaches incorporates the foundation of the Cognitive Sciences to introduce \textit{meaning} in the representations of image features.


\subsection*{Prototype Theory} 

The Prototype Theory~\citep{rosch1978cognition,rosch1975family, rosch1988coherences,geeraerts2010theories,minda2002comparing,rosch1975cognitive,rosch1978principles,zaki2003prototype} analyzes the internal structure of categories and introduces the prototype-based concept of categorization. It proposes categories representation as heterogeneous and not discrete, where the features and category members do not have the same relevance within the category.  Rosch~\citep{rosch1975cognitive,rosch1978principles} obtained evidence that human beings store first the \textit{semantic meaning of category} based on the degrees of representativeness~(\textit{typicity}) of category members, and then its specificities.

\begin{figure}[t]
	\begin{center}
		\includegraphics[width=0.9\linewidth]{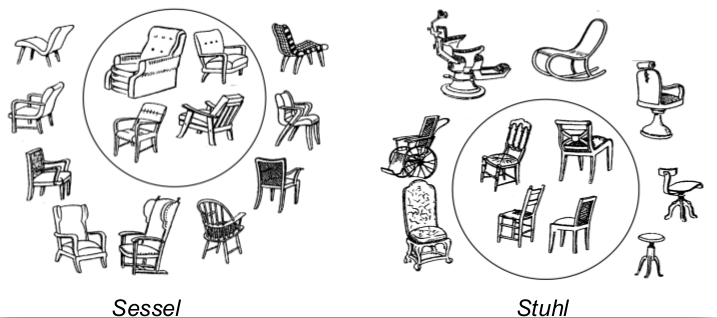}
	\end{center}
	\caption{\textit{Category prototypicality organization}. Figure shows the \textit{Sessel} and~\textit{Stuhl} experiment conducted by Gipper~(Figure adapted from~\cite{geeraerts2010theories}). That experiment studies the meaning of German words \textit{Stuhl}~(chair) and \textit{Sessel}~(comfortable chair) and shows that within the \textit{chair} category, category \textit{central semantic meaning} can change depending of observed feature relevance~(weights) and object typicality. This phenomenon is described in contemporary semantics as a \textit{prototypicality organization}~\citep{rosch1978principles, geeraerts2010theories} and constitutes one of the motivations of our proposal.} 
	\label{fig:chairs_experiment}
\end{figure}

The \textit{category prototype} was formally defined as the clear central members of a category~\citep{geeraerts2010theories,rosch1975cognitive,rosch1975family}. The attributes of these focal members are those that are structurally the most salient category properties, and conversely, a member occupies the focal position because it shows the most salient features of the category~\citep{rosch1975family,geeraerts2010theories}.
Rosch \citep{rosch1975cognitive,rosch1978principles,rosch1988coherences} showed that human beings store the category knowledge as a semantic organization around the category prototype~(\textit{prototypicality organization}).  Figure~\ref{fig:chairs_experiment} shows an example of the \textit{prototypicality organization} phenomenon \citep{rosch1978principles,rosch1978cognition,geeraerts2010theories}.
Finally, object categorization is obtained based on the similarity of a new exemplar with the learned categories prototypes \citep{rosch1978principles, rosch1988coherences}.

Rosch~\citep{rosch1975cognitive,rosch1978principles,rosch1978cognition,rosch1988coherences}  showed the important of making distinctions between various phenomena that may be associated with \textit{prototypicality}. For \cite{geeraerts2010theories} the concept
of prototypicality is in itself a prototypically clustered one for four characteristics in which the concepts of \textit{non-discreteness} and \textit{non-equality} (either on the \textit{intensional} or on the \textit{extensional} level) play a major distinctive role. Four characteristics are frequently mentioned as typical of prototypicality  in prototypical categories~\citep{rosch1975cognitive,rosch1978principles,geeraerts2010theories}: \textit{i)} categories exhibit degrees of typicality; not every member is equally representative in the category~(\textit{extensional non-equality}); \textit{ii)} categories are blurred at the edges~(\textit{extensional non-discreteness}); \textit{iii)} categories are clustering into \textit{family resemblance structure}; that is, the category semantic structure takes the form of a radial set of clustered and overlapping members~(\textit{intensional non-equality}); and \textit{iv)} categories cannot be defined by means of a single set of criteria~(necessary and sufficient) attributes~(\textit{intensional non-discreteness}). The \textit{prototypicality effects}~(see Table~\ref{table:prototypicality-effects}) surmise the importance of the distinction between central and peripheral meaning of the object categories~\citep{geeraerts2010theories}. 


\begin{table}[t]
	\caption{Two-dimensional conceptual map of prototypicality effects~\citep{geeraerts2010theories}.}
	\label{table:prototypicality-effects}
	\resizebox{\columnwidth}{!}{%
		\begin{tabular}{|l|c|c|}
			\hline
			
			& \textit{extensional} & \textit{intensional}\\
			\hline
			\hline
			\textit{non-equality} & Difference of typicality&Clustering into family\\{\scriptsize(salience effect,}& and
			membership salience & resemblances\\ {\scriptsize core/periphery)}&&\\
			\hline
			\textit{non-discreteness} & Fuzziness at the edges,&Absence of necessary and\\{\scriptsize (demarcation }& membership uncertainty& sufficient definitions\\ {\scriptsize problems, flexibly)}&&\\
			\hline
		\end{tabular}
	}
\end{table}

\section{Computational Prototype Model}
\label{sec:CPM_model}

\cite{rosch1975cognitive, rosch1978principles} showed that human beings learn the central semantic meaning of categories~(the prototype) and include it in their cognitive processes. Based on these assumptions, our object semantic description approach follows the flow of conceptual processes presented in Figure~\ref{fig:motivation} as a hypothesis for simulating the human behavior in object features description. 
Since our proposal requires as \textit{priori knowledge} the prototypes representation of objects categories, we need a procedure to represent the prototype of a specific category.

\subsection{Semantic Representation}
Category semantic structure~(\ie, \textit{central and peripheral meaning}) is related with differences of typicality and membership salience of category members~(\textit{extensional non-equality}).
The prototype can be understood as the ``average'' of the abstractions of all objects in the category \citep{sternberg2016cognitive}; it summarizes the most representative members~(or features) of the category.
The combination between observed object features and features relevance for the category enables the grouping of objects into family resemblance~(\textit{intensional non-equality}). This approach justifies the object's position within the semantic structure of the category and allows typical objects to be grouped into the semantic center of the category~(\textit{prototypical organization}).

Let $ O $ be an \textit{universe of objects};  
$ C = \left\{  {c_1, c_2,...,c_n}\right\} $ be the finite set of objects categories labels that partition $ O $;  $O_{c_i} = \left\{{o \in O: {\small category(o)} = c_i}\right\}$ is the set of objects that share the same \textit{i}-th category $c_i \in C$, $\forall i = 1,...,n;$ and $F =\left\{ {f_1, f_2,...,f_m}\right\} $ be a finite set of distinguishing features of an object.

\begin{definition}{\textit{Semantic prototype.}}
	We call the \textit{central meaning} of the category  $c_i \in C$, \textit{semantic prototype} of $c_i$-category, or simply \textit{semantic prototype}, to the ``average'' and  standard deviation of each  features of all \textit{typical objects} within the  $c_i$-category, along with a measure of the relevance of those features. Formally, our semantic prototype is a $3$-\textit{tuple} $ P_i = \left( {M_{i},\Sigma_{i},\Omega_{i}}\right) $ where $\:  \forall i = 1,...,n;\forall j = 1,...,m $: 
	\begin{enumerate}[i)]
		\item $ M_i = \left[  {\mu_{i1}, \mu_{i2},...,\mu_{im}}\right] $ is a nonempty $m$-dimensional vector, where $\mu_{ij}$ is the \textit{mean} of j-$th$ feature of features extracted for \textit{only typical objects} of $c_i$-category; 
		
		\item $ \Sigma_{i} = \left[ {\sigma_{i1}, \sigma_{i2},...,\sigma_{im}}\right] $ is a nonempty $m$-dimensional vector, where $\sigma_{ij}$ is the standard deviation of j-$th$ feature of features extracted for \textit{only typical objects} of $c_i$-category; 
		
		\item $ \Omega_{i} = \left[ {\omega_{i1}, \omega_{i2},...,\omega_{im}}\right] $ is a nonempty $m$-dimensional vector, where $\omega_{ij}$ is the relevance value of j-$th$ feature for the category $c_i \in C$.
	\end{enumerate}
	
	
	\label{def:semantic_pttype}
\end{definition}

\begin{definition}{\textit{Abstract prototype.}}
	The abstract semantic center of $ i $-th category $c_i \in C$,
	most prototypical element of  $i$-th category, ideal element of $i$-th category, or simply the \textit{abstract prototype} of $i$-th category, is the $ m $-dimensional vector $ M_i \in P_i = \left( {M_{i},\Sigma_{i},\Omega_{i}}\right)$ composed of the expected value of each features extracted for \textit{only typical objects} of $c_i$-category.
	\label{def:abstract_pttype}
\end{definition}

\subsection{Semantic Distance}

Our description approach (see processes 4 - 5 in Figure~\ref{fig:motivation}) needs a distance measure to compute the discrepancy between object features and category-typical features (semantic prototype). The distance metrics $ L1 $ and $ L2 $ could be good options if it did not assume that all object features have the same relevance.

According to the Prototype Theory: \textit{i)} each object feature has a relative relevance in the category and \textit{ii)} the relevance (or salience) of each category member is in accordance with the number and type of features present in the object. This approach can establish a degree of prototypicality of a specific element within the category~(\textit{extensional non-equality}). 

Some formal models of Experimental Psychology such as \textit{prototype model} \citep{reed1972pattern, homa1976category}, \textit{Multiplicative Prototype Model} (MPM) \citep{minda2001prototypes,minda2002comparing} and \textit{Generalized Context Model} (GCM) \citep{medin1978context, estes1986memory, nosofsky1986attention, zaki2003prototype}  proposed measures of semantic distances between stimulus that correspond to Prototype Theory foundations. Consequently, we generalized some of these semantics measures to propose a \textit{semantic distance metric}~(or dissimilarity function) that measures the discrepancy between two objects images~(or between an object image and its semantic prototype) based on observed features. 

\begin{definition}{\textit{Distance between objects.}}
	Let $ {o_1, o_2 \in O_{c_i}}$ be a representative objects of $ i $-th category $c_i \in C;$ $\emph{F}_{o_1}, \emph{F}_{o_2}$ the features of objects $o_1, o_2$ respectively. We defined the~\textit{objects distance} between $o_1$ and $o_2$ as the semantic distance given by:
	\begin{eqnarray} 
		\delta(o_1,o_2) = \sum_{j=1}^{m} \left| \omega_{ij}\right|\left|f_j^1-f_j^2\right|,
		\label{eq:objects distance}
	\end{eqnarray} 
	where $\omega_{ij} \in \Omega_{i},$ $f_j^1 \in F_{o_1}\,$ and $f_j^2 \in F_{o_2}\,, \: \forall i= 1...n;\,\forall j = 1...m$.
	\label{def:obj_distance}
\end{definition}

It is worth noting that our semantic \textit{distance between objects} is a generalization of \textit{the psychological distance between two stimuli} proposed in GCM formal model. Unlike the original formal \textit{Context Model}~\citep{medin1978context}, we assume that: \textit{i)} object features (stimuli) are not binary values~($f_j \in \mathbb{R}$); \textit{ii)} relevance~($\omega_{ij}$) (or cost of attention) of each $j$-th unitary object feature is forced to be strictly positive, but has no upper limit $(\sum_{j=1}^{m} \omega_{ij}\neq 1)$.  We removed these constraints of GCM Model in order to model \textit{object features} and \textit{object features relevance} using the features and weights learned by classification models.


\begin{definition}{\textit{Prototypical distance.}}
	Let $ {\emph{o} \in O_{c_i}}$ a representative object of $ i $-th category $c_i \in C$, $\emph{F}_o$ the features of object $\emph{o}$ and  $ P_i = \left( {M_{i},\Sigma_{i},\Omega_{i}}\right)$ the semantic prototype of $c_i$-category. We defined as \textit{prototypical distance}  between $\emph{o}$ and $P_i$ the semantic distance:
	
	\begin{eqnarray} 
		\delta(\emph{o},P_i) = \sum_{j=1}^{m} \left| \omega_{ij}\right|\left|f_j-\mu_{ij}\right|,
		\label{eq:pttype_distance}
	\end{eqnarray} 
	
	where $\omega_{ij} \in \Omega_{i},\,\mu_{ij} \in M_{i},$ and $f_j \in F_o\,;$ $ M_{i},\Omega_{i} \in P_i$ $\: \forall i= 1...n;\,\forall j = 1...m$.
	\label{def:pttype_distance}
\end{definition}

Our prototypical distance is a generalization of semantic distance of MPM formal model~\citep{minda2001prototypes,minda2002comparing}.  Different from MPM model assumptions, we assumed that prototype features are not features of a real member of $ i $-th category, but features of expected  ideal member~(our \textit{abstract prototype}) of $ i $-th category ($ M_{i} \in P_i$).

\begin{definition}{\textit{Features metric space.}}
	Let ${F}_{c_i}$ be a non empty set of all objects features of category $c_i \in C$. Since the distance function $\delta : {F}_{c_i} \times {F}_{c_i} \to \mathbb{R}^{+}$ satisfies the axioms of \textit{non-negativity}, \textit{identity of indiscernible}, \textit{symmetry} and \textit{triangle inequality};~$\delta $ is a \textit{metric} in the features domain ${F}_{c_i}$. Consequently, ~$({F}_{c_i},\delta)$ is a \textit{metric space} or \textit{features metric space}.
	\label{def:metric_space}
\end{definition}

\begin{proof} Let $ {o_1,\: o_2,\: o_3 \in O_{c_i}}$ objects members of i-\textit{th} category~($c_i \in C$); $ F_1,\: F_2,\: F_3$ the corresponding object features with $ f_j^1 \in {F}_{1},\: f_j^2 \in {F}_{2},\: f_j^3 \in {F}_{3}$; $\forall i = 1,...,n;\forall j = 1,...,m $.  
	
	\begin{itemize}
		\item $\delta(o_1,o_2) \ge 0$~\textit{(non-negativity)}. \\ Since all terms in Equation~\ref{eq:objects distance} are non negative~($\ge 0$), $\delta(o_1,o_2) \ge 0$ by definition;  
		
		\item $\delta(o_1,o_2) = 0 \Leftrightarrow o_1 = o_2$ ~\textit{(identity of indiscernible)}.\\ 
		\begin{itemize}
			\item $\delta(o_1,o_2) = 0 \rightarrow o_1 = o_2$. \\ If $\delta(o_1,o_2) = 0$ then $\sum_{j=1}^{m} \left| \omega_{ij}\right|\left|f_j^1-f_j^2\right| = 0$; consequently, since all terms in Equation~\ref{eq:objects distance} are non negative, the above expression is true if each element in the sum is zero. Then, $\forall \left| \omega_{ij}\right|$ $ \neq 0,$ $\left|f_j^1-f_j^2\right| = 0 \rightarrow f_j^1=f_j^2 \rightarrow o_1 = o_2$; \\
			\item $o_1 = o_2 \rightarrow \delta(o_1,o_2) = 0 $. \\ If $o_1 = o_2 \rightarrow f_j^1=f_j^2 \rightarrow \left|f_j^1-f_j^2\right| = 0, \forall j = 1...m;$ then $ \delta(o_1,o_2) = 0$;
		\end{itemize}
		\item $\delta(o_1,o_2) = \delta(o_2,o_1)$ ~\textit{(symmetry)}.\\
		$\delta(o_1,o_2)$=$\sum_{j=1}^{m} \left| \omega_{ij}\right|\left|f_j^1-f_j^2\right|$ $= \sum_{j=1}^{m} \left| \omega_{ij}\right|$ $\left|f_j^2 - f_j^1\right|$ $ = \delta(o_2,o_1)$; 
		
		\item $\delta(o_1,o_3) \le \delta(o_1,o_2) + \delta(o_2,o_3)$~\textit{(triangle inequality)}.\\
		$\delta(o_1,o_2) + \delta(o_2,o_3)$ $ = \sum_{j=1}^{m} \left| \omega_{ij}\right|\left|f_j^1-f_j^2\right| +$ $\sum_{j=1}^{m}$ $\left| \omega_{ij}\right|$ $\left|f_j^2-f_j^3\right| = \sum_{j=1}^{m} \left| \omega_{ij}\right|(\left|f_j^1-f_j^2\right| + \left|f_j^2-f_j^3\right|)$ and by absolute value property $\left|f_j^1-f_j^2\right| + \left|f_j^2-f_j^3\right|$ $ \ge \left|f_j^1-f_j^3\right|$, then $\delta(o_1,o_2) + \delta(o_2,o_3) \ge \delta(o_1,o_3)$.
	\end{itemize}
\end{proof}

Also, note that if $E \subseteq {O}_{c_i}$ is a subset of $i$-th category, $\delta(E) = \sum \delta(\emph{o},P_i), \, \forall \emph{o} \in E$. Consequently, our prototypical distance satisfies the following properties:  \textit{i)}\textit{ null empty set}: ${\displaystyle \delta(\varnothing)=0}$; ~\textit{ii)} \textit{countable additivity}: for all countable collections ${\displaystyle \{E_{k}\}_{k=1}^{\infty }}$ of pairwise disjoint sets in $E$, $
{\displaystyle \delta \left(\bigcup _{k=1}^{\infty }E_{k}\right)=\sum _{k=1}^{\infty }\delta (E_{k})}$~(this property is easy to prove using mathematical induction).

\begin{corollary}
	The prototypical distance function from ${F}_{c_i}$ to the extended real number line, $\delta : {F}_{c_i} \to \mathbb{R}^{+}$, is a measure. Consequently, $({F}_{c_i},\delta)$ is a measurable space.
\end{corollary}

Since our statement of $({F}_{c_i},\delta)$ is a measurable space, we can use the generalization  of Chebyshev's inequality~\citep{chebyshev1867} to define the boundary of our semantic prototype representation.~\cite{chebyshev1867} asserted that the probability that a scalar random variable $ \xi $ with distribution $ \Pr $ differs from its mean $ \mu \in \mathbb{R} $  by more than $ \lambda \in \mathbb{R} > 0 $ standard deviations $ \sigma \in \mathbb{R} > 0 $ satisfies the relation:~
$\Pr(|\xi-\mu|\geq \lambda \sigma) \leq \min (1, \frac{1}{\lambda^2})$. 

\cite{saw1084chebyshev} and \cite{stellato2017multivariate} approached the problem of formulating an empirical Chebyshev inequality given $N$ i.i.d samples from an unknown distribution $\Pr$, and their empirical mean $\mu_{N}$ and empirical standard deviation $\sigma_{N}$. \cite{saw1084chebyshev} and \cite{stellato2017multivariate} derives a Chebyshev inequality bound with respect to the $ (N + 1) $-th sample. The Multivariate Chebyshev inequality \citep{stellato2017multivariate} can define the boundary for an ellipsoidal set centered at the mean.

\begin{definition}{\textit{Semantic prototype edges.}}
	Let $({F}_{c_i},\delta)$ be the metric space of object features of $ i $-th category $c_i \in C$. Let $E \subseteq {F}_{c_i}$ be a set of features extracted for \textit{only typical objects} of $c_i$-category, $N = \left|{E}\right|$, and $F_{o}$ the features of a object $o \in O_{c_i}$. We \textit{weakly} defined as \textit{edges} of our semantic prototype $ P_i = \left( {M_{i},\Sigma_{i},\Omega_{i}}\right)$,  the threshold vector $ \vv{\lambda_{i}} = \left[ {\lambda_{i1}, \lambda_{i2},...,\lambda_{im}}\right]$ that meets the expression:
	\begin{eqnarray} 
		\Pr(|f_j-\mu_{ij}|\geq \lambda_{ij} \sigma_{ij}) \leq \min (1, \frac{1}{\lambda_{ij}^2}) ,
		\label{eq:semantic_edges}
	\end{eqnarray} 
	
	where $f_j \in F_{o}$, $\mu_{ij} \in M_{i}$ and $\sigma_{ij} \in \Sigma_{i}$,$ \: \forall i= 1...n;\,\forall j = 1...m.$. Finally, given a probability    bound, it is possible to compute a threshold vector $ \vv{\lambda_{i}}$ and construct a confidence ellipsoidal set from the sample mean and covariance of only typical objects samples~(a stronger \textit{semantic prototype edges} definition can be performed using -- completely-- the ~\cite{stellato2017multivariate} statements).
	\label{def:semantic_edges}
\end{definition}

\begin{figure}[t]
	\begin{center}
		\includegraphics[width=0.9\linewidth]{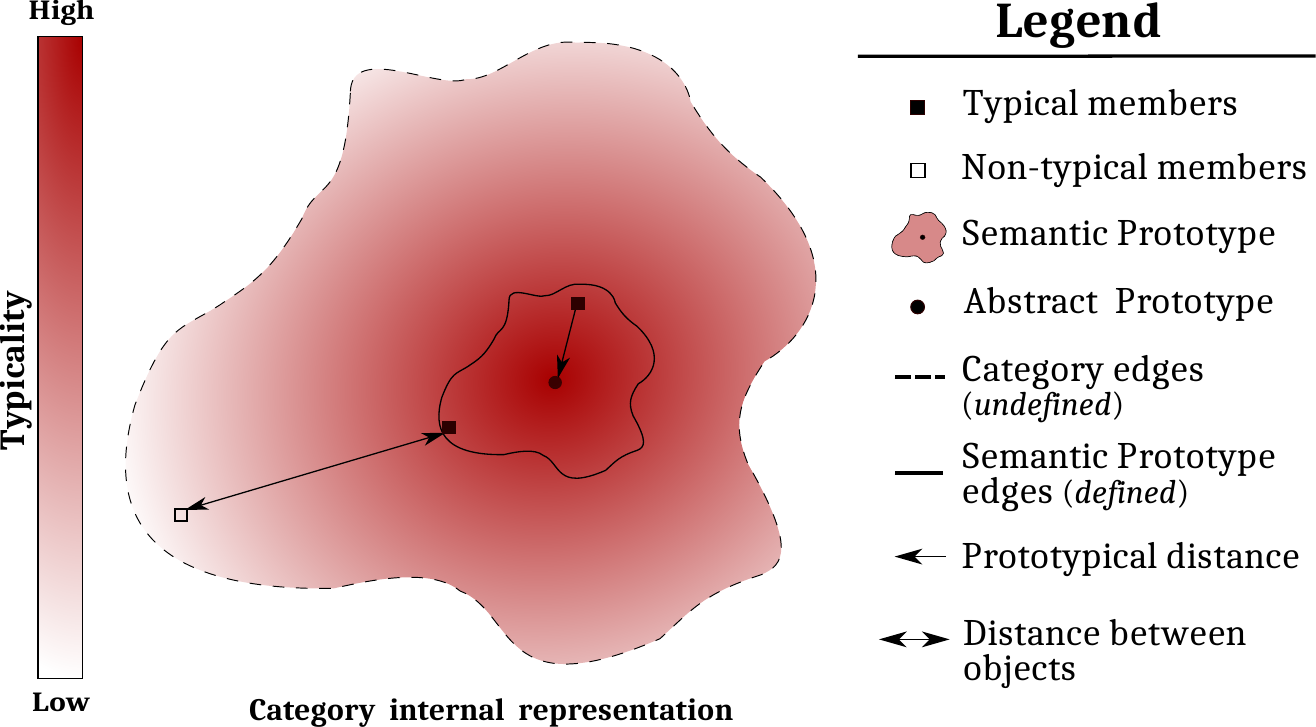}
	\end{center}
	\caption{\textit{Category internal structure}. Figure shows our expected semantic representation of category internal structure. Also we show the principal definitions of our Computational Prototype Model.}
	\label{fig:our_prototype_model}
\end{figure}

Figure~\ref{fig:our_prototype_model} shows the expected representation of category internal structure based on our Computational Prototype Model~(CPM)~[\textit{Semantic prototype}~(Definitions~\ref{def:semantic_pttype} and \ref{def:abstract_pttype}) + \textit{Semantic distance}~(Definitions~\ref{def:obj_distance} and \ref{def:pttype_distance})]. With our CPM model, we try to respect some important concepts of the Prototype Theory: \textit{i)} category prototype edges are defined with our vector $ \Sigma_{i} \in P_i = \left( {M_{i},\Sigma_{i},\Omega_{i}}\right)$; \textit{ii)} category edges are blurred~(not sharp defined) because our semantic prototype is not computed with all category elements~(only with typical elements); \textit{iii)} objects representativeness~(typicality) within the category is simulated with our prototypical distance.

\begin{figure*}[t]
	\begin{center}
		\includegraphics[width=0.88\linewidth]{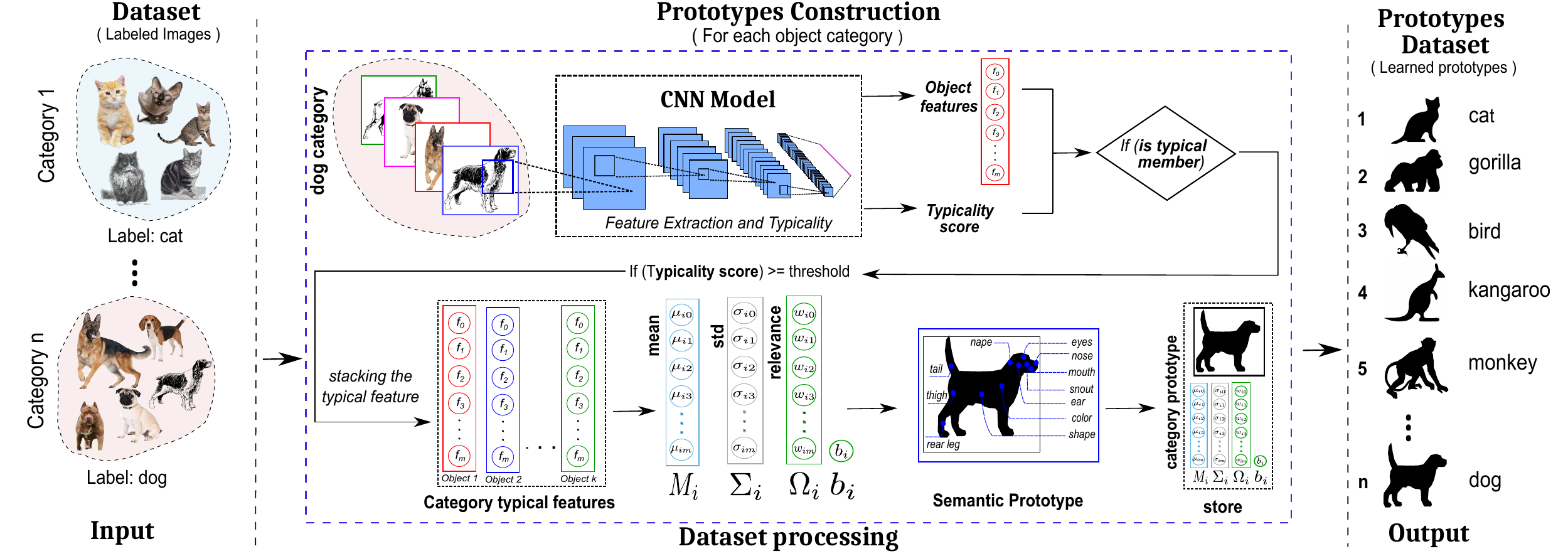}
	\end{center}
	\caption{\textit{Off-line construction of the semantic prototypes dataset}. Given a labeled images dataset, for each objects category present in the dataset, we compute our semantic prototype representation using Algorithm~\ref{alg:prototype}.}
	\label{fig:prototype_construction}
	
\end{figure*}

\subsection{Prototype Construction}
\label{sec:prototype}

Our \textit{semantic prototype} representation can be easily computed by any model with the ability to extract object features of images~($ F_o $) and learn the unitary relevance value~($\omega_{ij}$) of each j-$th$ object feature in i-$th$ category. We have also considered the elements typicality within the category to compute our \textit{semantic prototype}. Consequently, we need objects datasets with annotations of objects typicality scores.


Moreover, our object description approach presented in Figure~\ref{fig:motivation} attempts --- following the human behavior --- to use the same features extracted to classify and describe objects. First, we need to recognize the category to which the object belongs and then, find what the object features that distinguish it from others within the category are. However, how to model a global object description with similar behavior of the Figure~\ref{fig:motivation} diagram?


To address these issues, we rely on the fact that CNNs provide outstanding performance in image semantic processing and classification tasks. We used CNN -classification models for features extraction, recognition, and classification of the visual information received as input~(processes 1 to 4 in Figure~\ref{fig:motivation}). CNN-models, analogous to the human memory~\citep{fuster1997network}, make associations that keep the knowledge in its connection structures. Our method downloads that knowledge of pre-trained CNN-models into a semantic structure~(\textit{semantic prototype}), which aims is to stand for the central semantic meaning of learned categories~(see step 5 in Figure~\ref{fig:motivation}).

\begin{definition}{\textit{Convolutional semantic prototype.}}
	The \textit{convolutional semantic prototype} of $ i $-th category $c_i \in C$ is a $4$-tuple $ P_i = \left( {M_{i},\Sigma_{i},\Omega_{i},b_{i}}\right)$, where $M_{i},\Sigma_{i}$ are computed using features of $c_i$-category extracted from the \textit{fully convolutional layer} of pre-trained CNN - classification models; and $\Omega_{i},b_{i}$ are the learned parameters~(learned features relevance) of \textit{i}-$th$ category in the softmax layer. Next, we refer to \textit{convolutional semantic prototype} of the category as a \textit{semantic prototype}.
\end{definition}

\begin{algorithm}[t]
	\caption{Prototype Construction}
	\label{alg:prototype}
	\begin{algorithmic}
		\BState \emph{\textbf{Input}}: CNN-model $\Lambda$, objects dataset $O$, category $c_i$
		\BState \emph{\textbf{Output}}: Category Prototype ($P_i$)
		
		\State $ O_{c_i} \gets \left\{{o \in O: category(o) = c_i}\right\}$
		\State $ features\_block \gets \left\{ \right\}$
		\For {$ o \in O_{c_i}$ }
		\If {$o$~\textit{is\_typical}}
		\State $ F_{o} \gets \Lambda.features\_of(o)$
		\State $ features\_block \gets features\_block \cup F_{o} $
		\EndIf        
		\EndFor
		\State $ \Omega_{i},b_{i} \gets \Lambda.sofmax\_weight\_learned\_of(c_i)$
		\State $ M_{i},\Sigma_{i} \gets compute\_stats(features\_block)$
		\State \Return $ \left( {M_{i},\Sigma_{i},\Omega_{i},b_{i}}\right) $    
	\end{algorithmic}
	
\end{algorithm}

Algorithm~\ref{alg:prototype} details the computation of a \textit{semantic prototype} for a specific category. Given a labeled object images dataset, for each object category in dataset, we use Algorithm~\ref{alg:prototype} to compute the correspond semantic prototype~(\textit{off-line processing}). The resulting \textit{semantic prototypes dataset} is used as \textit{prior knowledge} in our prototype-based description model~(see Figure~\ref{fig:motivation}). Figure~\ref{fig:prototype_construction} shows the main steps and concepts of our prototype construction algorithm.

\subsubsection*{Semantic prototype visualization}

Visual representation of semantic prototypes allows presenting a visual summary of category typical features. Some approaches~\citep{wohlhart2013optimizing, li2018deep} learn prototypes representations in the image domain; and consequently, prototype visualization is simply the image learned. These approaches require a considerable computational expense to learn its prototypes visualization.~\cite{wohlhart2013optimizing} introduced the learning of image prototypes representations in the back-propagation process. On the flip side, \cite{li2018deep} used an encoder-decoder deep architecture to learn prototypes visualization. Since our semantic prototype representation is constructed straightforwardly from pre-trained CNN classification models, the methods mentioned above are not appropriate to visualize our prototype representation.

\begin{figure}[b]
	\begin{center}
		\includegraphics[scale=0.95,width=0.98\linewidth]{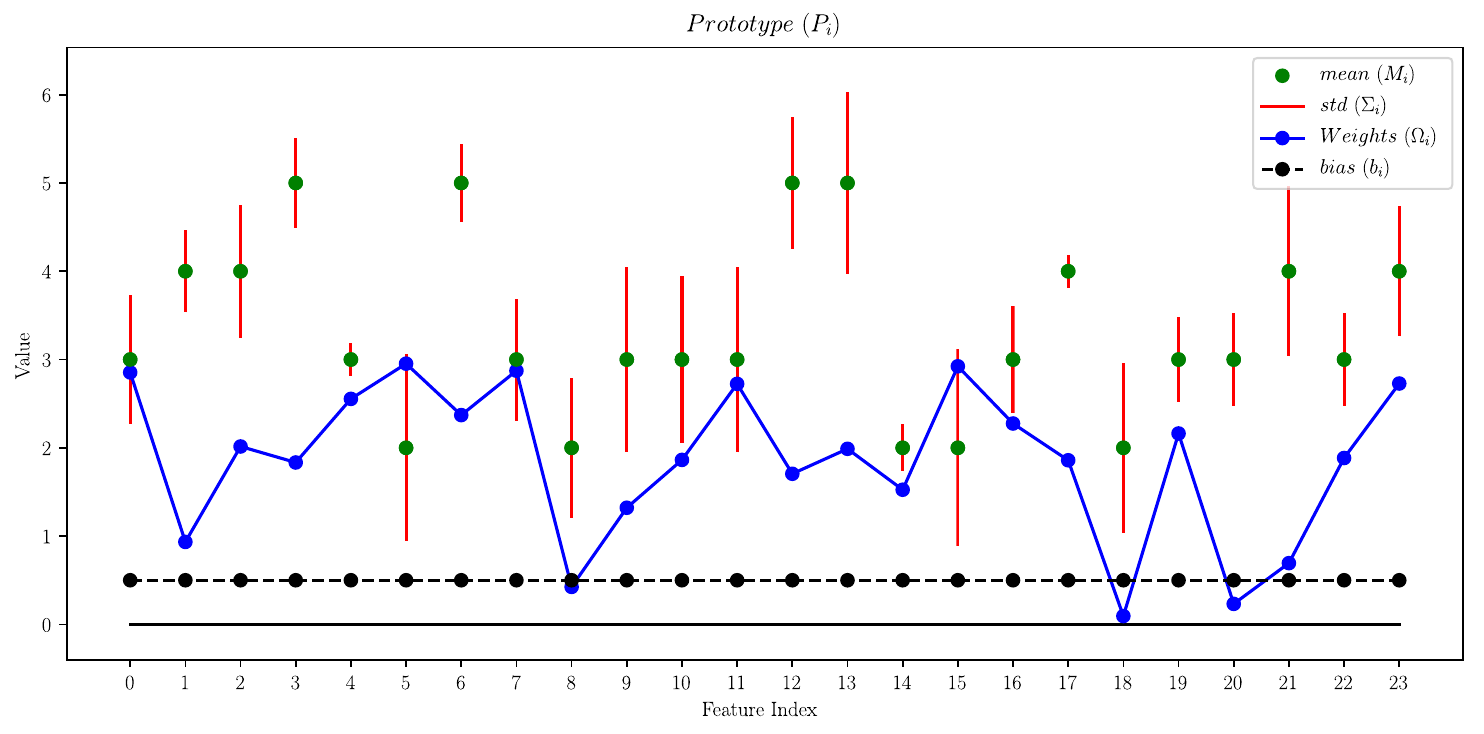}
	\end{center}
	\caption{Visualization of our semantic prototype representation  $P_i = \left( {M_{i},\Sigma_{i},\Omega_{i},b_{i}}\right)$ of $i$-th category. We showed the \textit{m}-dimensional vector $M_{i}$ (mean of\textit{ typical} members features) and \textit{m-}dimensional vector $\Omega_{i}$ (measure of features relevance within $i$-th category) in \textit{green} and~\textit{blue} colors, respectively. The \textit{m}-dimensional vector $ \Sigma_{i}$~(standard deviation of \textit{typical objects} features of $i$-th category) is represented as feature boundary~(in red lines) for each $j$-th unitary feature. Learned bias value $b_{i}$ is represented as a \textit{m}-dimensional vector.} 
	\label{fig:graphic_prototype}
\end{figure}

\cite{binder2016toward} proposed a circular visualization of semantic  mean attribute vectors  for concrete object noun categories.  Consequently, a simple approach for visualizing our semantic prototype representation is to visualize the distribution values of each $ m $-dimensional vector that compose the $P_i$-tuple of our semantic prototype definition. 

Figure~\ref{fig:graphic_prototype} shows an illustration of our semantic prototype representation corresponding to $ i $-th category. We showed each tuple-member ($ m $-dimensional vector) that composes the proposed semantic prototype of $i$-th category, $ P_i = \left( {M_{i},\Sigma_{i},\Omega_{i},b_{i}}\right)$. We represented the learned bias value $ b_{i}$ as the bias $ m $-dimensional vector $ \vv{b_i} \in \mathbb{R}^{m}, \, \vv{b_i} = \dfrac{b_{i}}{m} \cdot \vv{1}$ such that $b_i = \sum_m \vv{b_i}$.

It is noteworthy that our semantic prototype has a values distribution that is characteristic of $ i $-th  category it represents. \Ie, our \textit{semantic prototype} can be understood as a DNA chain that stands for the category members' typical features. The semantic prototype representation uniqueness is guaranteed by the relevance vector~($\Omega_{i}$), which was learned specifically for that $ i $-th category when the CNN-classification model was trained.

\section{Global Semantic Descriptor}
\label{sec:descriptor}

\begin{figure*}[t]
	\begin{center}
		\includegraphics[scale=0.5,width=0.98\linewidth]{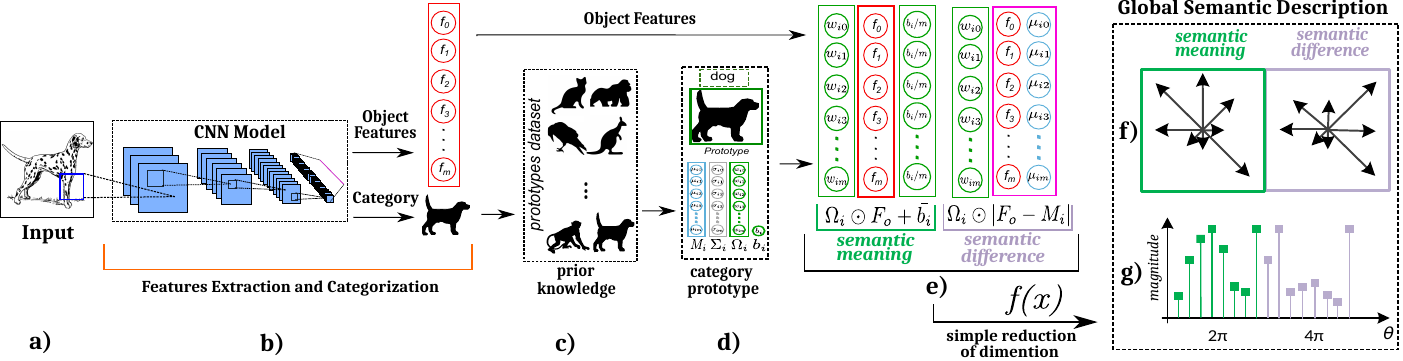}    
	\end{center}
	\caption{\textit{Overview of our prototype-based description model.} Set of steps to transform the visual information received as input into a Global Semantic Descriptor signature.~\textit{a}) input image;~\textit{b)} features extraction and classification using a CNN-classification model;~\textit{c)} prototypes dataset;~\textit{d)} category prototype selection;~\textit{e)} global semantic description of object using category prototype;~\textit{f)} graphic representation of our Global Semantic Descriptor signature resulting from the dimensionality reduction function~($f(x)$); and ~\textit{g)} Global Semantic Descriptor signature.}
	\label{fig:methodology}
\end{figure*}

In the previous section, we presented a framework to encapsulate the central meaning~(semantic prototype) of an object category. In this section, we present how to introduce that semantic prototype representation to simulate the object semantic description work-flow depicted in Figure~\ref{fig:motivation}.

\subsection{Semantic Meaning}

Some cognitive neuroscience researches have studied the effect of \textit{semantic meaning} in object recognition task \citep{tulving2007coding,martin2007representation, collins2013conceptual}. When an object has been previously associated with some type of semantic meaning in the brain, people are more prone to identify the object \citep{tulving2007coding,martin2007representation} correctly.  Studies \citep{tulving2007coding,martin2007representation,collins2013conceptual} have shown that semantic associations allow a much faster recognition of an object, even when the task of object recognition becomes increasingly difficult (varying points of view, occlusion) \citep{collins2013conceptual}. Therefore, semantic associations based on object semantic meaning allow for faster object recognition.

Moreover, the fact that some CNN models (\eg, ResNet \citep{he2016deep}) outperform the human-reported performance ($5.1\%$\citep{ILSVRC15}) on large-scale visual object classification tasks, generated some cognitive studies \citep{yamins2014performance,cadieu2014deep,khaligh2014deep,cichy2017dynamics} to research the possible links between CNN models and visual system in the human brain. \cite{cichy2017dynamics} suggested that deep neural networks perform spatial arrangement representations like those performed by a human being.~\cite{khaligh2014deep} concluded that the weighted combination of features in the last fully connected layer of CNN models could thoroughly explain the inferior temporal cortex in the human brain. We lay hold of these theoretical foundations to model our representation of \textit{objects semantic meaning}.



\begin{definition}{\textit{Semantic value.}}  Let be $F_o$ observed features of an object ${\emph{o} \in O}$ ($F_o= \left\{ f_{1}, f_{2},...,f_{m}\right\}$).
	The \textit{semantic meaning} of object features $F_o$ for category $c_i \in C$, \textit{summary value} of features $F_o$, or simply \textit{semantic value} of $F_o$ in $c_i$-category is an abstract value:
	$z= \sum_{m} \omega_{ij}f_j + b_{i},$ 
	where $\omega_{ij} \in \Omega_{i},\,f_{j} \in F_o $. Consequently, the semantic value of ideal member of $c_i$-category,  \textit{central semantic meaning} of $c_i$-category or \textit{summary value} of the semantic prototype $P_i= \left( {M_{i},\Sigma_{i},\Omega_{i},b_{i}}\right)$ is the \textit{semantic value} $\hat{z_{i}} = \sum_{m} \omega_{ij}\mu_j + b_{i},$ where $\omega_{ij} \in \Omega_{i},$ and $\mu_{ij} \in M_{i}$ are the \textit{abstract prototype} features, $\forall i= 1,...,n;\,\forall j = 1,...,m.$ 
	\label{def:semantic_value}   
\end{definition}

Note that our \textit{object semantic value} is exactly the same value used to object categorization in softmax layer of CNN-classification models. Hence, our approach of object semantic description based on prototypes assumes as object \textit{semantic meaning vector}, the semantic vector~($ \vv{z} = \Omega_{i}  \odot F_{o} + \vv{b_i}$) constructed with the element-wise operations to compute the object \textit{semantic value}~(Definition~\ref{def:semantic_value}). Our \textit{semantic meaning} representation uses a bias vector~($ \vv{b_i}$) to uniformly dissolve the bias value in each semantic vector component~($b_i = \sum_m \vv{b_i})$. With this approach it is enough a sum of each \textit{semantic meaning vector} component to recover the \textit{object semantic value}~($z = \sum_m \vv{z})$.
Accordingly, our \textit{semantic meaning vector} contains the same semantic definitions used for CNN models to categorize an object within a specific category.

\subsection{Semantic Difference}
\label{subsect:semantic_distance}


We stand for the \textit{semantic distinctiveness} of an object for specific $c_i$-category as the semantic discrepancy between object features and features of the most prototypical~(ideal) element of $c_i$-category~(abstract prototype of $c_i$-category). Since object features~($F_o$) and \textit{abstract prototype} of $c_i$-category~($M_i \in P_i$) belong to the same features domain~(features metric space), we apply our \textit{prototypical distance} as measure of the objects distinctiveness within a category. 

Consequently, our approach assumes as object \textit{semantic distinctiveness vector}, the \textit{semantic difference vector}~($\vv{\delta} = \Omega_{i} \odot \left|F_{o} - M_{i}\right|$) constructed with the element-wise operations to compute the object \textit{prototypical distance}~(Definition~\ref{def:pttype_distance}). Our~\textit{semantic difference vector} is the weighted~($\Omega_{i}$) \textit{residual vector}~($ \vv{r} = \left|F_{o} - M_{i}\right|$) composed of absolute values of the difference between each object feature and each feature of $c_i$-category abstract prototype.

Note that our \textit{object semantic difference}~(or our prototypical distance) can be understood as the sum of absolute difference between the \textit{object semantic meaning vector}~($\vv{z}$) and the \textit{central semantic meaning vector}~($\vv{\hat{z_i}}$) of $c_i$-category. Thus, Equation~\ref{eq:pttype_distance} is equivalent to $\sum_{j=1}^{m} \left|\vv{z_j} -\vv{\hat{z_{ij}}}\right|$ = $\sum_{j=1}^{m} \left|\omega_{ij}f_j-\omega_{ij}\mu_{ij}\right|$ = $\delta(\emph{o},P_i) $ when $\forall \omega_{ij} \in \Omega_{i}$, $\omega_{ij} \geq 0$ (we introduced this  $\omega_{ij}$ constraint in the semantic distance of MPM model). Therefore, our \textit{object semantic difference} representation has the advantage that elements vector sum is enough to retrieve the object \textit{prototypical distance}~($\delta = \sum_m \vv{\delta})$.


Figure~\ref{fig:methodology} depicts an overview of our prototype-based description model. Our \textit{Global Semantic Descriptor based on Prototypes} \textit{(GSDP)} uses as a requirement the \textit{prior knowledge}  of each category prototype~(prototypes are precomputed off-line using Algorithm~\ref{alg:prototype}). After feature extraction and categorization processes (Figure~\ref{fig:methodology}b), we use the corresponding category prototype for semantic description of object features. We show in Figure~\ref{fig:methodology}e) 
the steps to introduce the category prototype into the global semantic description of object's features. A drawback of our object semantic representation~(Figure~\ref{fig:methodology}e) is having high dimensionality, since it is based on  \textit{semantic meaning vector}~($\vv{z}$) and \textit{semantic difference vector}~($\vv{\delta} = \Omega_{i} \odot \vv{r}$). The large dimensionality of our feature vectors might make its use unfeasible in common computer vision tasks 
\citep{han2017scnet,kim2017fcss}. Figure~\ref{fig:methodology} and Algorithm~\ref{alg:global_descriptor} detail the main steps of our approach; note that steps follow the same work-flow of human description hypotheses depicted in Figure~\ref{fig:motivation}.


\begin{algorithm}[t]
	\caption{Global Semantic Descriptor $\psi$}
	\label{alg:global_descriptor}
	\begin{algorithmic}[1]
		\BState \emph{\textbf{Input}}: Image of an object $ o $ 
		
		\BState \emph{\textbf{Output}}: Object semantic signature~($\psi_o$)
		\BState \emph{\textbf{Prior Data}}: Trained CNN-model $\Lambda$, $prototypes\_dataset$
		\State $ F_{o}, c_{i} \gets \Lambda.features\_and\_prediction(o)$
		\State $ {M_{i},\Sigma_{i},\Omega_{i},b_{i}} \gets prototypes\_dataset(c_i)$
		\State $ \textit{meaning} \gets f\left({F_o,\Omega_{i},b_{i},\textit{meaning}}\right)$
		\State $ \textit{difference} \gets f\left({\left|F_{o} - M_{i}\right|,\Omega_{i},b_{i},\textit{distinctiveness}}\right)$
		\State \Return $ \textit{meaning} \oplus \textit{difference} $
		
	\end{algorithmic}
\end{algorithm}

\subsection{Dimensionality Reduction}

\begin{figure*} [t!]
	\begin{center}
		\includegraphics[scale=0.5, width=0.98\linewidth]{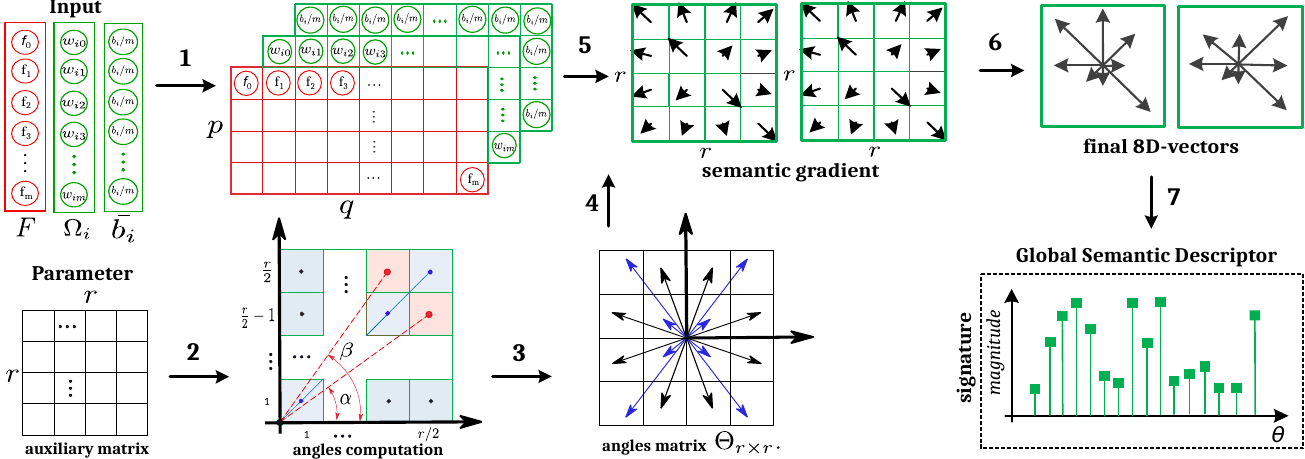}
	\end{center}
	\caption{\textit{Dimensionality reduction function}.
		Figure shows our transformation $f(x)$ to convert the high dimensionality of our object semantic representation into the corresponding semantic descriptor signature. Final signature is constructed by concatenating each  8D-vector computed from each unitary semantic gradient. We showed the trivial case when the input m-dimensional vectors have 2 times auxiliary matrix dimension~($m=p \cdot q  \: and \: p=r;\: q =2r$); consequently,  output signature has 2 times (16D) the 8D-vector dimension.
	}    
	\label{fig:dim_reduction}
\end{figure*}

Several dimensionality reduction algorithms such as PCA \citep{abdi2010principal} and NMF \citep{lee2001algorithms} are based on discarding features that do not generate a meaningful variation. Although these approaches work on some tasks, after applying these algorithms, we lost the ability of data interpretation~\citep{abdi2010principal}. From the Prototypes Theory perspective, discarding features is no suitable when it is applied to the semantic space due to the absence of necessary and sufficient definitions to categorize an object (\textit{intensional non-discreteness}). Occasionally when discarding features might lead in discarding elements of the category \citep{geeraerts2010theories}. For instance, there may be some objects within the category that do not have some category typical features~(flying is a typical feature of bird category; however, a penguin is a bird that does not fly).



We proposed a simple transformation function~$f(x)$ to compress our global semantic representation of the object's features~(Figure~\ref{fig:methodology}e) in a low dimensional global semantic signature~(Figure~\ref{fig:methodology}g).
Our transformation function aims to reduce our semantic representation dimensionality while keeping the property of \textit{easy retrieve} the \textit{object semantic meaning} and \textit{object semantic difference} from the final descriptor signature. Our final descriptor signature~$(\psi)$ is computed by concatenating the corresponding signatures of \textit{semantic meaning vector}~($\vv{z}$) and \textit{semantic difference vector}~($\vv{\delta}$) compressed with our $f(x)$ transformation~(see Algorithm~\ref{alg:global_descriptor}).


Figure~\ref{fig:dim_reduction} shows the main steps of our $f(x)$ transformation. We use a square auxiliary matrix~($\chi_{r\times r}$) as a parameter to control the descriptor signature dimensionality. The auxiliary matrix dimensions is a parameter that allows us to control the final GSDP-signature size, \ie; larger auxiliary matrix dimensionality leads to smaller GSDP signature; and vice versa. 

The main steps showed in Figure~\ref{fig:dim_reduction} can be summarized as: \textbf{1)} Resize the input vectors in the best 2D dimensional configuration of matrices~($ p \times q $) whose dimensions are multiples of $ r $~(auxiliary matrix dimension). \textbf{2-3)} Compute the angles matrix ($\Theta_{r\times r}$) with angles formed by the position of each feature with respect to auxiliary matrix $\chi_{r\times r}$ center; to achieve uniqueness the diagonal angles were evenly distributed between the magnitudes of angles $ \alpha $ and $ \beta$.~\textbf{4) - 5)} Create \textit{unitary semantic gradient} for each auxiliary matrix mapped within $p \times q$ matrices; each~\textit{semantic gradient} is constructed using the \textit{angle matrix} ($\Theta_{r\times r}$), and \textit{magnitude} and \textit{sign} of semantic vectors computed using Definition~\ref{eq:pttype_distance} and~\ref{def:semantic_value}.~\textbf{6)} Reduce the semantic gradient to 8-vectors similarly to SIFT approach~\citep{lowe2004SIFT};~\textbf{7)} Concatenate, for each auxiliary matrix $\chi_{r\times r}$ mapped, the corresponding unitary 8D-signatures resulted of flow $4$-$6$. Algorithm~\ref{alg:dim_reduction} details all steps.

\begin{algorithm}[t]
	\caption{Dimensionality Reduction $f(x)$}
	\label{alg:dim_reduction}
	\begin{algorithmic}[1]
		\BState \emph{\textbf{Input}}:
		\textit{m-dimensional vector}~$\alpha$, $\Omega_{i}, b_{i},$ \textit{type}
		\BState \emph{\textbf{Output}}: Semantic signature
		\BState \emph{\textbf{Parameter}}: Auxiliary matrix~$\chi_{r\times r}$
		
		\State $ \bar{b_i} \gets \frac{b_i}{m} \qquad \qquad$  { //~\footnotesize m-dimensional vector $ \bar{b_i}$ $(b_{i} = \sum_{m} \bar{b_i})$ }
		
		\State $ \chi_{r\times r} \gets shape(r,r) \: $ { //~\footnotesize setting auxiliary matrix dimension}  
		\State Computing angles matrix: $\Theta_{r\times r} = angles\_from( \chi_{r\times r})$
		
		Find new shape p,q from k
		\State Finding the optimal configuration $p, q$ where $p \equiv 0\ (\textrm{mod}\ r)$, $q \equiv 0\ (\textrm{mod}\ r)$ and $ m = p \cdot q$
		
		\State $ \alpha, \Omega_{i},\bar{b_{i}} = reshape\_to\_matrix_{p\times q}(\alpha,\Omega_{i},\bar{b_i})$
		
		\State $ signature \gets \left [  \right ]$
		\For {$ j = 1,...,\frac{p}{r}; k = 1,...,\frac{q}{r} $}
		
		\State Mapping $ \chi_{r\times r}^{jk}$ in $ \alpha, \Omega_{i},\bar{b_i} $
		\State Computing $ \vv{z_i}^{jk} $ using \textit{Hadamard product}  $\odot $.
		\State $ \vv{z_i}^{jk} = 
		\begin{cases}
		\Omega_{i}^{jk} \odot \alpha^{jk} + \bar{b_i}^{jk},&{\text{if}~\textit{type} = \textit{meaning}}\\
		\left|\Omega_{i}^{jk}\right| \odot \alpha^{jk},& \textit{otherwise}
		\end{cases}  $
		\State $g\ ^{jk} \gets vectors(\Theta_{r\times r},\left|\vv{z_i}^{jk}\right|, sign(\vv{z_i}^{jk}))$.
		
		\State $signature\ ^{jk}(l) =\sum g^{jk}(\theta), \forall \theta \in \Theta_{r\times r} : \theta_l -45 < \theta \leq \theta_l $ with $ \theta_l = l  \cdot \frac{\pi}{4}, \forall l = 1,...,8 $    
		\State $ signature \gets signature \oplus signature\ ^{jk} $
		\EndFor
		\BState \Return $signature $    
	\end{algorithmic}
\end{algorithm}

Hence, our final descriptor signature preserves the \textit{object semantic meaning}~(Property~\ref{property:semantic_value}) and the \textit{object semantic difference} (Property \ref{property:object_distance}) presented in our first global semantic representation of object features (Figure \ref{fig:methodology}e). Additionally, depending of the input vector, our descriptor can uses $f(x)$ transformation to construct global semantic representations (signatures) with different meanings within $i$-th category (Property \ref{property:polymorphism}). In other words, our descriptor can construct semantic representations (see Figure \ref{fig:methodology}e) for: \textit{i)} an object, \textit{ii)} ideal category member (abstract prototype), and \textit{iii)} category semantic meaning  encapsulated with semantic prototype boundaries.


\subsection{Descriptor Properties}

\begin{property}{\textit{Semantic meaning preservation.}} The semantic descriptor signature preserves the \textit{object semantic value}:~$\sum_{l=0}^{\left | \psi  \right |/2} \psi[l] = \hat{z}.$ 
	\label{property:semantic_value}  
\end{property} 
\begin{proof}
	To prove this, it suffices to follow backward through steps~$6$ and $ \left [9, 17 \right]$ of Algorithm~\ref{alg:global_descriptor} and \ref{alg:dim_reduction}, respectively.  $\sum_{l=0}^{\left | \psi  \right |/2} \psi$ = $\sum f\left({\alpha,\Omega_{i},b_{i},\textit{meaning}}\right)$ = $\sum \sum_j \sum_k g^{jk}$ = $\sum \Omega_{i} \odot \alpha + \bar{b_i}$= $\sum \vv{z}$ = $\hat{z};~\alpha \in \left\{{M_i, F_o}\right\}.$    
\end{proof}

\begin{property}{\textit{Prototypical distance preservation.}}~
	If ${\emph{o} \in O_{c_i}}$ is a object of $i$-th category, the object signature $\psi_o$ preserves the object \textit{prototypical distance}:~$\sum_{l=\left | \psi_o  \right |/2}^{\left | \psi_o  \right |} \psi_o[l] = \delta(\emph{o},P_i).$ 
	\label{property:object_distance}
\end{property}
\begin{proof}
	Similar to the previous proof, but using distinctiveness vector~$(\left|\Omega_{i}\right| \odot \left|F_{o} - M_{i}\right|)$ through steps~$7$ and $ \left [9, 17 \right]$ of Algorithm~\ref{alg:global_descriptor} and \ref{alg:dim_reduction}, respectively.\\
	$\sum_{l={\left | \psi  \right |/2}}^{\left | \psi  \right |}$  = $\sum f({\left|F_{o} - M_{i}\right|,\Omega_{i},b_{i},\textit{distinctiveness}})$
	= \\$\sum \sum_j \sum_k  g^{jk}$ = $\sum \left|\Omega_{i}\right| \odot \left|F_{o} - M_{i}\right|$ = $\sum \vv{\delta}$ = $\delta(\emph{o},P_i).$ 
\end{proof}

\begin{property}{\textit{Structural polymorphism.}}~Our global semantic descriptor GSDP has the polymorphic property of describing, with the same structural representation, distinctly different semantic meanings within the $c_i$-category. Thus, our descriptor uses the category prototype $P_i = \left({M_{i},\Sigma_{i},\Omega_{i},b_{i}}\right)$ to construct different semantic signature taxonomies: 
	\begin{enumerate}[i)]
		\item an object~$o\in O_{c_i}$,~$\psi_o$ = $\psi(F_{o},\left|F_{o} - M_{i}\right|,\Omega_{i},b_{i})$ = \\ $ f\left({F_{o},\Omega_{i},b_{i},\textit{meaning}}\right)$ $ \oplus$ \\$ f(\left|F_{o} - M_{i}\right|,\Omega_{i},b_{i},\textit{distinctiveness})$;
		\label{property:polymorphism_object}
		\item \textit{central semantic meaning} of $i$-th category~(abstract prototype),~$\psi_{P_i}$ = $\psi(M_{i},\left|M_{i} - M_{i}\right|,\Omega_{i},b_{i})$ \\= $\psi(M_{i},\vv{0},\Omega_{i},b_{i}) $; \label{property:polymorphism_prototype}
		\item \textit{semantic meaning} of $i$-th category~(semantic prototype), $\psi_i$ = $\psi(M_{i},\Sigma_{i},\Omega_{i},b_{i})$.
		\label{property:polymorphism_category}
	\end{enumerate}
	\label{property:polymorphism}  
\end{property}

\section{Experimental Evaluation}

\subsection{Experimental Setup}
Aside from performing experiments using benchmark image datasets with fixed-size, size-normalized and centered images like MNIST \citep{lecun1998} and CIFAR \citep{krizhevsky2010convolutional}, we also evaluated our approach on ImageNet \citep{ILSVRC15} as real images dataset. For each image dataset, we used a CNN-classification model for feature extraction and classification (see Figure~\ref{fig:methodology}b). Thus, we used a CNN-MNIST and CNN-CIFAR models based on \textit{LeNet} \citep{lecun1998} and \textit{Deep Belief Network} \citep{krizhevsky2010convolutional} architectures for image classification in MNIST and CIFAR datasets, respectively. Also, we conducted experiments in ImageNet using VGG16 \citep{simonyan2014very} and ResNet50 \citep{he2016deep} models as background of our global semantic description model. Note that our \textit{prototype-based description model} depicted in Figure~\ref{fig:methodology}, is scalable and can easily be adapted to any other CNN-classification model.

\subsubsection*{Prototypes Dataset Construction}

Our \textit{prototype-based description model} requires \textit{prototypes dataset} as category \textit{prior knowledge} (see Figure \ref{fig:methodology}c) to build object semantic representations (see Figure \ref{fig:methodology}e) that stand for the object distinctiveness within the category. In the experiments, we computed prototypes datasets with CNN-MNIST, CNN-CIFAR, VGG16, and ResNet50 models in MNIST, CIFAR, and ImageNet datasets, respectively. 


For feature extraction, we assumed as object features those extracted from the last dense layer (before the softmax layer) of the CNN-model.  Our approach needs typical objects of categories or any information about typicality score~(or typicality degree) of objects belonging a specific category to build the proposed semantic prototype properly.  However, none of the images datasets used have this annotation. \cite{lake2015deep} showed that the output of the last layer of CNN models could be used as a signal for how typical is an input image.  Consequently, we used as \textit{typicality score} of objects the strength of classification response to the category of interest. Specifically, we assumed as typical members of a category those elements that are --- unequivocally --- classified as category members (\textit{typicality score} $>0.99$) by CNN models (see Figure \ref{fig:prototype_construction}). Finally, for each category in datasets, we extracted features of typical members and computed the correspond \textit{semantic prototype} (see Definition~\ref{def:semantic_pttype})  using  Algorithm~\ref{alg:prototype}.

\subsection{The Semantics behind our Computational Prototype Model}
Achieving the member's prototypical behavior within a category is one of the motivations and theoretical basis of our approach. Nevertheless, there is no defined metric to quantify whether our representation correctly captures the category semantic meaning. This lack of a metric is a consequence of the fact that there is no defined metric to evaluate the object typicality level within a category robustly; this skill is still reserved only for human beings.

In this section, we analyzed the semantics captured by our CPM Model~(\textit{semantic prototype} + \textit{prototypical distance}). The CPM model pursues two main goals: \textit{i)}~capture, with the semantic prototype, the central semantic meaning of a specific object category; \textit{ii)}~simulate, in a comparable way to  human beings, that visually typical elements of category are organized close~(based on our prototypical distance metric) to category prototype. Since we do not have annotated images with the object typicality score to robustly evaluate the semantic captured by our representation, we used another approach to analyze the semantics behind our CPM model.

\subsubsection{Central and Peripheral meaning} 

In this section, we analyzed the \textit{central and peripheral} meaning captured by our CPM model. Since we defined \textit{abstract prototype} as the \textit{abstract semantic center} of category, we observed how relevant (or visually representative) -- for the category -- are those elements allocated by our CPM model in \textit{center and periphery} of category. Expressly, this experiment aims to know what is the visual representativeness of category members closest and furthest from the category semantic center (our abstract prototype).

\begin{figure*}[]
	\begin{center}
		\includegraphics[width=0.87\linewidth]{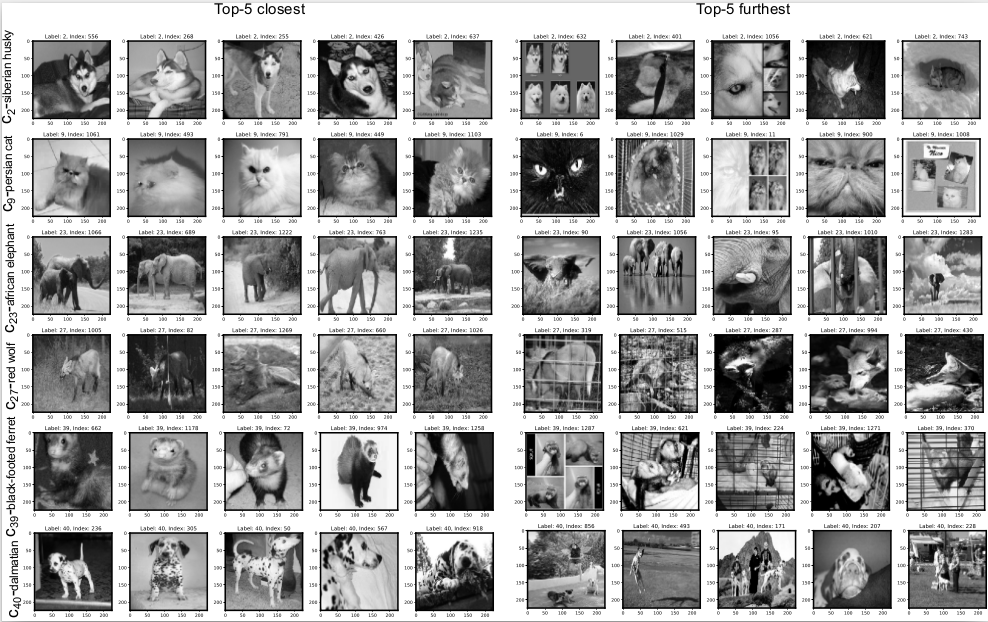}
	\end{center}
	\caption{(\textit{from left to right}) Top-5  most relevant members identified by our CPM model for a categories sample of ImageNet dataset.~(\textit{left}) Top-5 elements closest to semantic prototype of corresponding category; index value represents the element position within the category dataset.~(\textit{right}) Top-5 elements furthest from the semantic prototype of the category. Object image features was extracted with VGG16 model.}
	\label{fig:top5_ImageNet}    
\end{figure*}

\begin{figure}[]
	\begin{center}
		\includegraphics[width=0.98\linewidth]{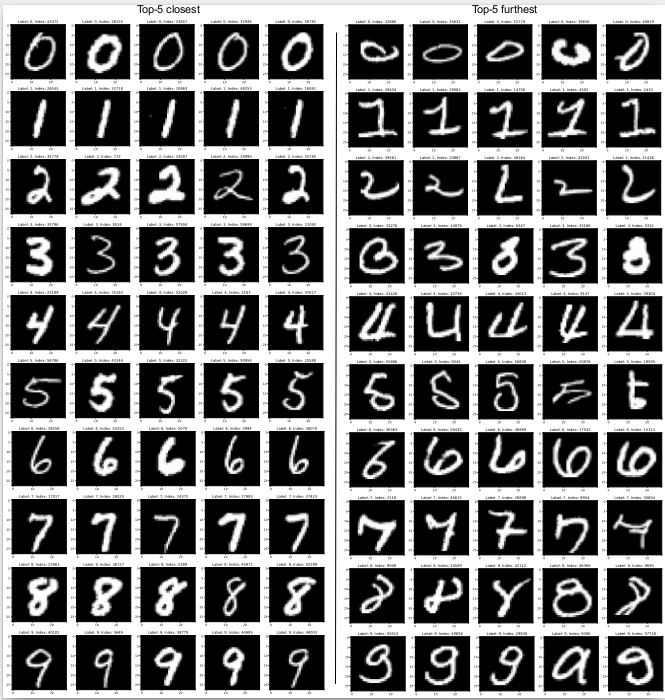}
	\end{center}
	\caption{(\textit{from left to right}) Top-5  most relevant members identified by our CPM model for all  MNIST dataset categories.~(\textit{left}) Top-5 elements closest to category abstract prototype of corresponding category;~(\textit{right}) Top-5 elements furthest from the category semantic prototype. Image features was extracted with CNN-MNIST model.}
	\label{fig:top5_mnist}    
\end{figure}

To achieve such a goal, we extracted object image features using a CNN model and computed our \textit{prototypical distance} for all members of $ i $-th category. Finally, the objects images are sorted in ascending order based on the prototypical distance value of each element. Figures \ref{fig:top5_ImageNet} and \ref{fig:top5_mnist} present some examples of central and peripheral meaning captured by our CPM model for images categories of ImageNet and MNIST datasets using VGG16 and CNN-MNIST models -- respectively -- as image feature extractors.

Notice that our proposal for object image semantic interpretation -- using our CPM model in CNN image feature domain -- attempts to assign a visual representativeness value (or typicality) to object image within the category to which it belongs.

Figure \ref{fig:top5_mnist} shows the \textit{Top-5 closest} and \textit{Top-5 furthest} elements from category center (abstract prototype) detected by our CPM model in MNIST categories. For instance, our proposal finds as \textit{typical} elements (Top-5 closest) of \textit{number three category} the handwritten digits with features that are, undoubtedly, distinctive of $c_3$-category. Our CPM model also can find the peripheral meaning of the category. Members with fewer characteristic features of \textit{number three}, or little readable, are placed in the periphery (Top-5 farthest) away from the \textit{central semantic meaning}, but keeping the category features~(it still belongs to the category). Similar to a human being, our CPM model can find the Top-5 farthest members of \textit{number three} category that are a number $3$, but not a typical number $3$.

\begin{figure*}[h]
	\begin{center}
		\includegraphics[scale=0.8,width=0.95\linewidth]{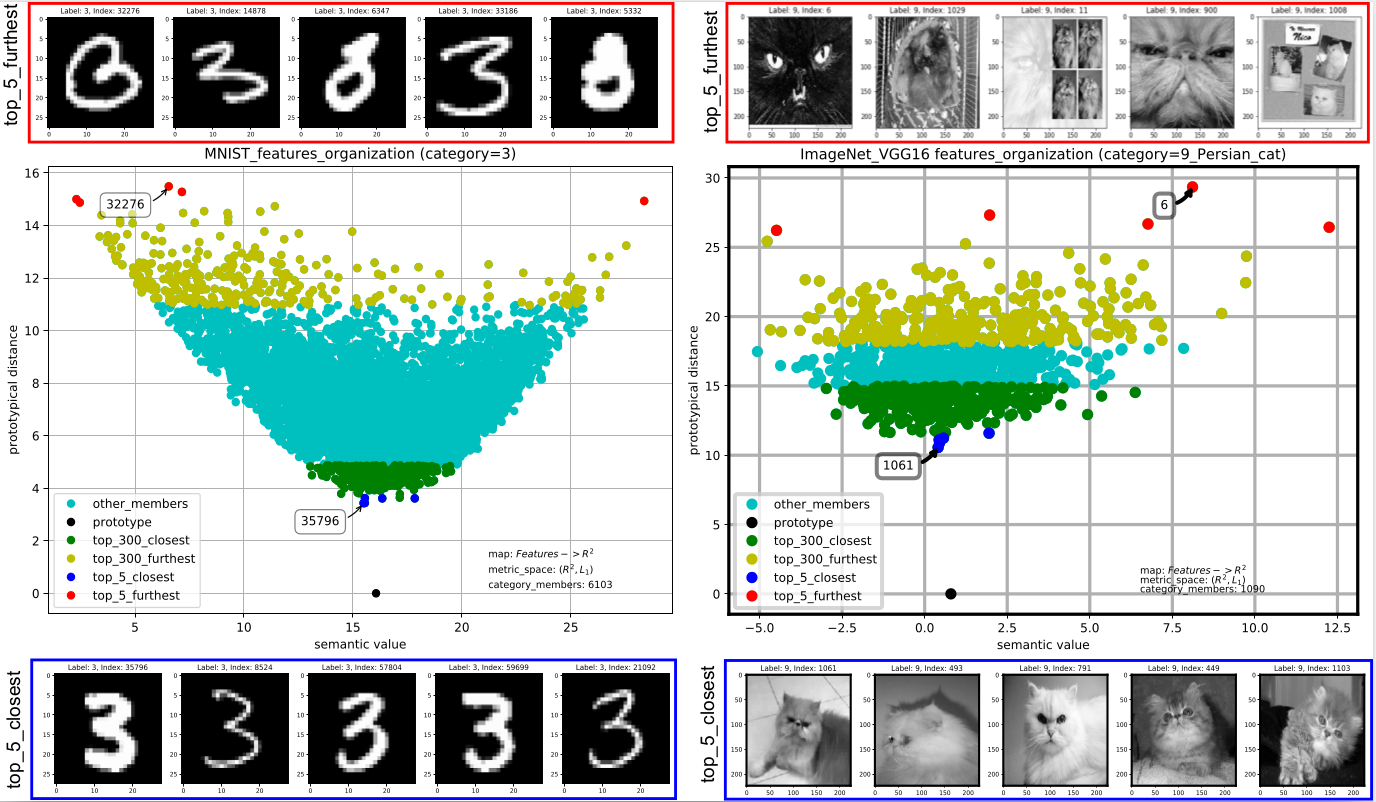}
		\caption{\textit{Prototypical organization within categories.} Figure shows the internal structure   of \textit{number three} and \textit{Persian cat} categories of MNIST and ImageNet datasets, respectively. We represented each category member  using image features extracted with CNN-MNIST and VGG16 models. We represented with color degrees  the category internal disposition respect its prototype. 
			In button and top, from left to right, the mapped Top-5 elements closest (in blue) and furthest (in red) to the mapped semantic prototype (in black) of each category. Image  dataset index of the first Top-5 element is annotated inside the black box.}
		\label{fig:prototypical_organization}
	\end{center}
\end{figure*}

Figure \ref{fig:top5_ImageNet} presents the semantic interpretation of visual image information performed by our CPM model in real object images of ImageNet dataset. Note that category members recognized by our CPM model as Top-5 closest members (left column) to category semantic center are easy recognized by human beings, as it exhibits the typical features of an object category. Also, we observed that Top-5 furthest elements (right column) from the semantic prototype (or less representative members of $ i $-th category) detected by CPM model although retaining some category features are not easily recognized by human beings. That is, our CPM model can identify the  most/least visually representative  category members, and --- correctly --- recognizes as category periphery members those elements, where not all typical category features are identified:  category typical colors, size, shape, etc. are not easily distinguishable; or object pose in the image does not exhibit these representative features of $ i $-th category.  The experiments performed allow to assume that our CPM model can capture the central/peripheral semantic meaning of images categories. But, we still need to answer to the question: Can our CPM model organize all category elements prototypically?




\subsubsection{Prototypical Organization} 

The experiments in this section aim to visualize the internal semantic structure of the category using the semantic meaning encapsulated by our CPM model for each category member. Based on features extracted from objects images, we analyzed the object prototypical behavior observing where it is positioned within the category by our CPM model~(using our \textit{prototypical distance}). Visualizing the semantic position of each category member with respect to \textit{category abstract prototype} constitutes a simple approach to see the internal semantic structure of the entire category structure. We need to corroborate that our CPM model can correctly interpret the object image features and position it semantically within the category, keeping a \textit{prototypical organization} of the category.

Note that our CPM model uses $ m $-dimensional object features from CNN image-features domain. Accordingly, visualizing the category's internal structure is infeasible in $ m $-dimensional features space since most techniques of data visualization are based on features discarding. From the perspective of the Prototype Theory foundations, features discarding approach can be problematic~(\textit{intensional non-discreteness}). For this reason, we used topology techniques to make object image interpretation based on all observed features. We constructed a map function to show that our CPM model can simulate the prototypical organization of members within a category.

Let $({F}_{c_i},\delta)$ and $(\mathbb{R}^{2},L_1)$ be  \textit{metric spaces}~(see Definition~\ref{def:metric_space}), and $\rho$ a function that maps image object features to $(\mathbb{R}^{2},l_1)$ metric space using its \textit{semantic value} and its~\textit{prototypical distance}. \Ie,  $\rho: {F}_{c_i} \to \mathbb{R}^{2}\mid \rho(o\in O_{c_i})$ = $\rho(F_o)$ = $p(z_o,\delta(o,P_i))$, where $F_o$ are the object features, $z_o$ is the  \textit{object semantic value},~$\delta(o,P_i)$ is the \textit{object prototypical distance}; the point ~$p(x,y) \in \mathbb{R}^{2}$ and $L_1$ is L1-norm condition.

Let be the objects $o_1,o_2 \in O_{c_i}$, and $p_1$ = $\rho(o_1)$, $p_2$ = $\rho(o_2)$ be the corresponding mapped points in $(\mathbb{R}^{2},L_1)$ metric space. Then, the Sum of Absolute Difference (SAD) between $p_1$ and $p_2$ is $L_1(p_1, p_2)$ = $L_1(\rho(o_1),\rho(o_2))$ = $ L_1(p(z_1,\delta(o_1,P_i)),p(z_2,\delta(o_2,P_i)))$ = $|z_1 - z_2|$  + $|\delta_1 - \delta_2|$; using Definitions~\ref{def:obj_distance}, \ref{def:pttype_distance}, and~\ref{def:semantic_value} we have:~$\delta(o_1,o_2) \leq L_1(p_1,p_2) \leq 2 \delta(o_1,o_2)$. Consequently, for every $F_{o_1}, F_{o_2}$ $\in F_{c_i}$ and $ \varepsilon > 0$, exists a $ \varphi$ = $\frac{\varepsilon + 1}{2} > 0 $ such that: $\delta(o_1,o_2)<\varphi$ $\Rightarrow$ $ L_1(\rho(o_1),\rho(o_2)) < \varepsilon$, \ie,~$\rho $ \textit{is continuous}. This means that every element of  $\rho(o_1)$ neighborhood in $(\mathbb{R}^{2},L_1)$ metric space, also belongs into $o_1$ neighborhood  in $({F}_{c_i},\delta)$ metric space (if $\rho(o_1)$ = $p_1$, $ \forall p \in \left\{p_1 \: \textit{neighborhood}\right\}$, $\rho^-(p) \in \left\{o_1 \: \textit{neighborhood}\right\}$). Consequently, the observed behavior of $i$-th category internal structure -- in terms of distance metrics-- in $(\mathbb{R}^{2},L_1)$ metric space is equivalent to the behavior in feature metric space $({F}_{c_i},\delta)$.


Figure~\ref{fig:prototypical_organization} shows an example of the internal semantic structure of MNIST and ImageNet images categories mapped using $\rho$.  Note how Top-5 closest members (based on our \textit{prototypical distance}) are mapped~(\textit{in blue}) and positioned near (based on L1 distance) to the mapped abstract prototype~(\textit{in black}). The Top-5 most visually representative members of each category in $({F}_{c_i},\delta)$ metric space are the same Top-5 most representative (closest to mapped \textit{abstract prototype}) in $(\mathbb{R}^{2},L_1)$ metric space. Likewise, the Top-5 fewer representative members~(\textit{in red}) continue to be positioned in the category peripheries, far away from the category abstract prototype ~(our central semantic meaning representation). The experiments show a prototypical organization of mapped members within the category in $(\mathbb{R}^{2},L_1)$ metric space. Consequently, based on $\rho$ properties, a similar grouping of objects based on family resemblance is preserved in CNN-features metric space.

Our approach to visualize the category internal structure also allows observing other semantic phenomena related to the object image visual representativeness. The experiments showed that \textit{object semantic value} and \textit{object prototypical distance} place the object image in a unique semantic position within the category internal structure. This result shows that our approach of constructing a semantic object representation~(see Figure~\ref{fig:methodology}e) based on vector versions of \textit{semantic value} and \textit{prototypical distance} can be able to describe the object image semantically.


\begin{figure*}[]
	\begin{center}
		\includegraphics[scale=0.98,width=0.85\linewidth]{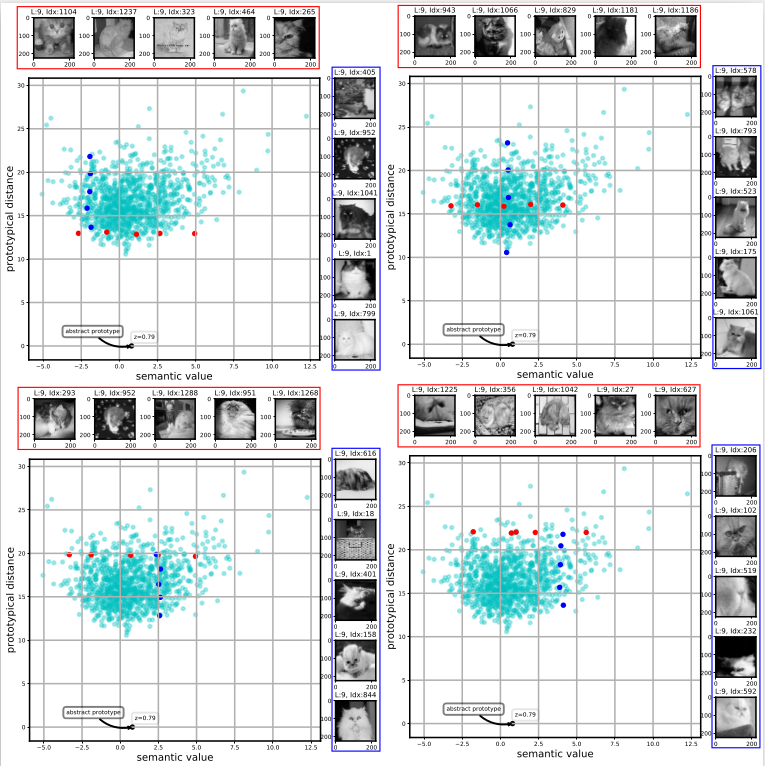}
		
		\caption{\textit{Typicality score analysis}. Objects images with same prototypical distance and different semantic values (in red) have similar visual representativeness within the category, and category members with different prototypical distance and same semantic value~(in blue) are visually different. Also, we observe that object image visual representativeness~(\textit{typicality}) decreases as prototypical distance increases. Object image features were extracted with VGG16 model.
		}
		\label{fig:semanticV_vs_prototypicalD}
	\end{center}
\end{figure*}




\subsubsection{Image Typicality Score}

We observed that the shape of category internal structure -- in $(\mathbb{R}^{2},L_1)$ metric space using our visualization approach -- strongly depends on semantic values distribution and prototypical distance distribution. Consequently, in this section we analyzed the relationship between \textit{semantic value} and \textit{prototypical distance} variables. Also, we examined how the variations of these variables can influence on object image visual representativeness (typicality) within the category. 

Our \textit{prototypical distance} can be understood as the semantic difference between the object semantic meaning  and the semantic meaning of category abstract-prototype (see Subsection \ref{subsect:semantic_distance}). Specifically, if features relevance~($ \Omega_{i} $) of $ i $-th category is strictly positive ($\omega_{ij} \geq 0$, $\forall \omega_{ij} \in \Omega_{i}$) then, the variables  \textit{prototypical distance} and \textit{semantic value}  are -- by construction -- strongly correlated. However, experiments in MNIST, CIFAR and ImageNet datasets with each corresponding CNN-model showed that there is a small strength of a linear association between those two variables~(Pearson coefficient values between $-0.3$ and $0.3$), but it does not conclude that we can generalize a behavioral pattern between  \textit{object semantic value} and \textit{prototypical distance}. This Pearson correlation result is consequence of the fact that the weights learned in softmax layer (our feature relevance $ \Omega_{i} $) of CNN models used for feature extraction (CNN-MNIST, CNN-CIFAR, VGG16 and ResNet50) are not strictly positive. Consequently, the \textit{semantic value} is not a strong measure because the addends in the equation can cancel each other out (see Definition \ref{def:semantic_value}), and elements with same semantic value does not imply that elements are equal ($ z_{o_1} = z_{o_1} \centernot\implies o_1 = o_2$).

\cite{lake2015deep} showed that \textit{semantic value} can be used as a signal for how typical an input image looks like. In contrast to Lake~\etal~results, our experiments with VGG16 and ResNet50 models in ImageNet dataset showed that using the \textit{semantic value} as object typicality score can be problematic because objects with same semantic value do not imply same image visual typicality. Figure~\ref{fig:prototypical_organization} shows an example of this phenomenon. In \textit{Persian cat} ImageNet category, the 5th element of Top-5 closest to category prototype (in blue) has a semantic value like 2nd position element of Top-5 furthest (in red)~(both images \textit{semantic values} are $\approx$ 2), but objects images are visually different. That is, the \textit{semantic value} could be a necessary condition to image typicality representation, but it is not enough. On the other hand, note how our \textit{prototypical distance} can capture the visual typicality difference between those two objects images.

We observed what is the image visual information behavior when one of those semantic variables (\textit{semantic value} and \textit{prototypical distance}) change. We kept constant the value of one variable, and then, we analyzed the visual representativeness of the corresponding object images when the value of another variable increase. Figure \ref{fig:semanticV_vs_prototypicalD} shows an example of this experiment within the Persian-cat ImageNet category. Note how for a fixed \textit{prototypical distance} (\textit{elements in red}), the semantic value variation does not generate significant changes in image visual representativeness (typicality) within the category. In contrast, for a fixed semantic value~(\textit{elements in blue}), the prototypical distance variation generates typicality ordered changes in the image's visual information. We observed that when prototypical distance increases, object image visual typicality decreases. In contrast, the experiments did not allow to generalize a behavior pattern between semantic value and image typicality.

Based on the results of our experiments, we assumed that our semantic prototype representation correctly captures the central semantic meaning of images categories. Even with different CNN models and images datasets, our CPM model organizes the internal category structure following a prototypical organization of category members. Besides, we showed that our~\textit{prototypical distance} influences elements arrangement around the category semantic prototype. Since our prototypical distance is a metric in CNN-feature domain, our semantic distance can be used as object image typicality score within the category~(\textit{typicality score} $(\emph{o}) = 1/ \delta(\emph{o},P_i)$). 

\subsection{Global Semantic Descriptor based on Prototypes}

\subsubsection{Descriptor Configuration}


By construction, the dimensionality of our GSDP descriptor signature depends of the object image CNN-features dimensionality (image features extracted with CNN classification model used in background), and dimensions of auxiliary matrix~($\chi_{r\times r}$) used as parameter in our $ f(x) $ transformation~(see Figure~\ref{fig:dim_reduction} and Algorithms~\ref{alg:global_descriptor}, \ref{alg:dim_reduction}). With higher auxiliary matrix dimensionality, smaller is our GSDP signature size; and vice versa.

Consequently, since we needed the size variation of image CNN-features to evaluate our prototype-based description model, we used different CNN models as images features extractors. CNN models selection criteria were based on trying to evaluate our semantic description approach in different contexts: image CNN-features with different sizes, CNN models with varied architecture and depth, and image datasets of diverse nature (image resolution, image type, etc.). Also, for each CNN model used, we configured (using the auxiliary-matrix parameter) our semantic descriptor to return GSDP-signatures with noticeably different dimensionality.


Table~\ref{table:descriptor_size} presents details of GSDP descriptor settings used to construct each semantic GSDP signature evaluated in our experiments. For each CNN-model used, we exhibit the CNN-feature status at each step of the workflow of our dimensionality reduction function~($f(x)$)~(see Figure~\ref{fig:dim_reduction}). Table~\ref{table:descriptor_size} shows the CNN classification models used as feature extractor; image CNN-feature length ($|F|$); new CNN-feature shape~($F_{p\times q}$) after apply Step 1 of our $f(x)$ transformation; auxiliary matrix dimension ($\chi_{r\times r}$) used as parameter (we used two different configurations for each CNN model); number of matrices that make up our semantic gradient~($g^{jk}$); dimensionality of intermediary feature constructed with our $f(x)$ transformation ($|f(x)|$); and final length of GSDP signature~($|\psi|$) for each descriptor setting. 


Note that our global semantic descriptor executes twice the $f(x)$ transformation to reduce the dimensionality of  \textit{semantic meaning} and \textit{semantic difference} representations (see Algorithm~\ref{alg:global_descriptor}). Consequently, GSDP signatures dimensionality is two times the dimensionality of $f(x)$ transformation features.

\begin{table}[t]
	\caption{Available GSDP descriptor signature dimensions for each CNN classification model used as features extractor.}
	\label{table:descriptor_size}
	\begin{center}
		\resizebox{\columnwidth}{!}{%
			\begin{tabular}{lllcccc}
				\hline
				\multicolumn{1}{c}{CNN Model}  & \multicolumn{1}{c}{ $|F|$} & \multicolumn{1}{c}{$F_{p\times q}$} & \multicolumn{1}{c}{$\chi_{r\times r}$} & \multicolumn{1}{c}{$ g^{jk} $} & \multicolumn{1}{c}{$|f(x)|$} & \multicolumn{1}{c}{$|\psi |$} \\ \hline \hline
				\multirow{2}{*}{CNN-MNIST} & \multirow{2}{*}{128}  & \multirow{2}{*}{${16\times 8}$} & $8\times 8$ & $2\times 1$ & 16 & 32 \\
				&       &                                                 & $4\times 4$   & $4\times 2$ &64  & 128  \\     \hline
				\multirow{2}{*}{CNN-CIFAR} & \multirow{2}{*}{512}  & \multirow{2}{*}{${32\times 16}$} & $8\times 8$   & $4\times 2$ &64  & 128  \\
				&                       &                                                              & $4\times 4$   & $8\times 4$ &256 & 512 \\     \hline
				\multirow{2}{*}{VGG16}     & \multirow{2}{*}{4096} & \multirow{2}{*}{${64\times 64}$}& $16\times 16$ & $4\times 4$ &128 & 256  \\     
				&                       &                                                              & $8\times 8$   & $8\times 8$ &512 & 1024  \\ \hline
				\multirow{2}{*}{ResNet50}  & \multirow{2}{*}{2048} & \multirow{2}{*}{${64\times 32}$}& $16\times 16$ & $4\times 2$ &64  & 128 \\
				&                       &                                                             & $8\times 8$   & $8\times 4$ &256 & 512 \\     \hline
			\end{tabular}
		}
	\end{center}
	\vspace*{-\baselineskip}
\end{table}

\subsubsection{Signature Semantic Information}  
\label{subsection:signature_information}
The experiments in the image CNN-features domain showed that \textit{object semantic value} and \textit{prototypical distance} organize all category members prototypically in a specific (and unique) position within the category semantic structure. The key idea behind our GSDP semantic descriptor is to encapsulate, in a vector representation, the same semantic interpretation --of image object features-- captured by our CPM model. In this section, we show that our GSDP descriptor encodes and preserves the semantic information contained in an object features~(semantic value and prototypical distance) used by our CPM model for semantic interpretation of object image. Also, we show how retrieving from GSDP descriptor signatures that semantic information and reconstructing the prototypical organization of object category achieved in the image CNN-features domain.


Let be $({\psi}_{c_i},L_1)$ the metric space of object descriptor signatures. Descriptor properties \ref{property:semantic_value} and \ref{property:object_distance} allow to easily recover the \textit{object semantic value} and \textit{prototypical distance} from GSDP descriptor signatures. 
Property \ref{property:polymorphism} enables us to build descriptor signatures for abstract prototypes of categories. Similarly to $\rho$ map, we can construct and show that map $\gamma:$ $({\psi}_{c_i},L_1)$ $\to(\mathbb{R}^{2},L_1) \mid \gamma(\psi_o \in \psi_{c_i})$ = $p(\sum_{0}^{\left | \psi \right |/2} \psi_o,\sum_{\left | \psi \right |/2}^{\left | \psi \right |} \psi_o)$ = $p(z_o,\delta(o,P_i))$ \textit{is continuous}. Hence, we can map all category descriptor signatures  to $(\mathbb{R}^{2}, L_1)$ metric space using $\gamma$ function.




With $\gamma$ map approach, we can reproduce the same semantic analysis performed in CNN-feature space. Experiments showed that the category prototypical organization achieved in $(\mathbb{R}^{2}, L_1)$ metric space is identical regardless of which $\gamma$ map (to descriptor signature domain) or $\rho$ map (to CNN-feature domain) function is used~(\eg~ Figure~\ref{fig:prototypical_organization} and \ref{fig:semanticV_vs_prototypicalD}). Consequently, the behavior observed in $(\mathbb{R}^{2}, L_1)$ metric space is equivalent to the behavior in feature metric space $({F}_{c_i},\delta)$ and descriptor signatures metric space~$({\psi}_{c_i},L_1)$. This means that our GSDP descriptor signature preserves, in its taxonomy, the same semantic information used by our CPM model to interpret object image CNN-features~(\textit{semantic value} and \textit{prototypical distance}).

\subsubsection{Signature Taxonomies}

\begin{figure}[b]
	\begin{center}
		\includegraphics[width=0.99\linewidth]{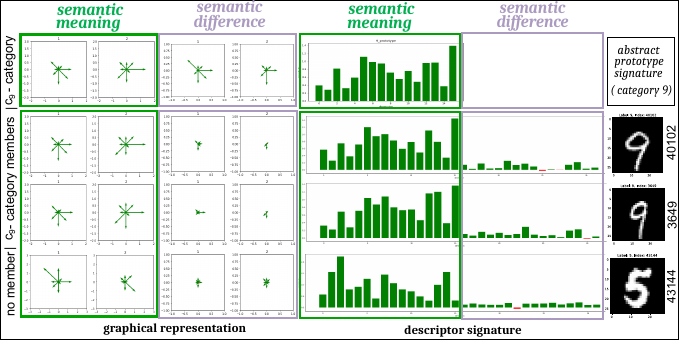}
	\end{center}
	\caption{\textit{Semantic signature taxonomies}. Figure shows an example of semantic signatures constructed with our GSDP descriptor for $c_9$-category in MNIST dataset. We show the abstract prototype signature, descriptor signatures examples of two $c_9$-category members and a member that does not belong to $c_9$-category. 
	}
	\label{fig:taxonomies_mnist}
	
\end{figure}

By definition, our GSDP descriptor uses category semantic prototypes as semantic distinctiveness generator of category members signatures. Elements with similar \textit{semantic meanings} and that sharing similar \textit{semantic differences} with the abstract prototype, will have similar GSDP semantic signatures. That is, since \textit{abstract prototype} can be understood as a DNA chain that stands for the typical CNN-features of category members, the \textit{abstract prototype signature} can be understood as a number distribution~(or smaller DNA chain signature) that stands for category members signatures.


Figure~\ref{fig:taxonomies_mnist} shows an example of the signatures taxonomies constructed with our GSDP semantic descriptor. We showed GSDP signatures constructed using CNN - MNIST model as features extractor of MNIST dataset images (signatures size = $32$ since we used the GSDP minimal setting (see Table \ref{table:descriptor_size})).
~Also, we showed the structural polymorphism property (Property~\ref{property:polymorphism}) of our GSDP descriptor to construct signatures for the \textit{central semantic meaning} (abstract prototype) and category members. With our approach, category members will have semantic signatures with a similar representation of category abstract prototype signature. Notice that very typical category elements will have descriptor signatures similar to the abstract prototype signature, and elements that do not belong to the category will have a quite different GSDP signature.

Our GSDP descriptor attempts to build -- using our semantic prototype representation -- a specific signature distribution for each object image category. Figure~\ref{fig:prototypical_organization} and \ref{fig:semanticV_vs_prototypicalD} show that category elements can be grouped, based on the meaning captured by our CPM model, by their family resemblance within the object category. However, this does not mean that in $ m $-dimensional image features space, there are no elements of other categories in the neighborhood of a specific element. Since t-SNE algorithm~\citep{maaten2008visualizing} can preserve the local structure, we used t-SNE to analyze the element neighborhood in $ m $-dimensional space.~\cite{maaten2008visualizing} exposed that points which are close to one another in the high-dimensional dataset will tend to be close to one another in the t-SNE low-dimensional map.

\begin{figure}
	\centering
	\subfigure[]
	{
		\includegraphics[width=1\linewidth]{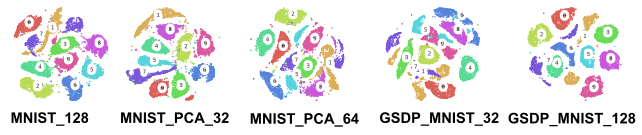}
		\label{fig:t-SNEmnist} 
	}
	\\
	\subfigure[]
	{
		\includegraphics[width=1\linewidth]{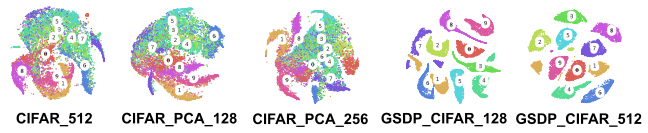}
		\label{fig:t-SNEcifar} 
	}
	\subfigure[]
	{
		\includegraphics[width=1\linewidth]{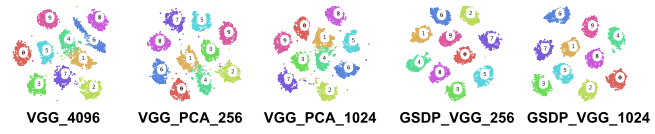}
		\label{fig:t-SNEvgg16} 
	}
	
	\subfigure[]
	{
		\includegraphics[width=1\linewidth]{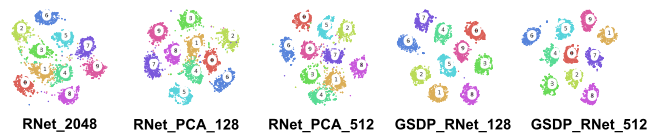}
		\label{fig:t-SNEresnet} 
	}
	\caption
	{
		\textit{t-SNE visualization.} \textit{a)} t-SNE visualization of features constructed with CNN-MNIST model in MNIST dataset; \textit{b)} t-SNE visualizations of features constructed with CNN-CIFAR model in CIFAR10 dataset; \textit{c,d)} t-SNE visualizations of first 10 categories of ImageNet dataset using features constructed with VGG16 and ResNet50 models, respectively. Each feature length was placed in the corresponding caption.
	}

	\label{fig:t-SNE}
\end{figure}


We analyzed the discriminative power and t-SNE visualization performance of our GSDP semantic image representation \textit{versus}  features extracted using CNN classification models. For each CNN-model used as background by our GSDP descriptor, we compared the t-SNE performance of features family built with each CNN model. We performed the t-SNE visualization experiment for features-family constituted by CNN features, corresponding GSDP semantic signatures, and reduced PCA versions of CNN-features (we reduced CNN-features to same GSDP feature dimensions). 

Figure \ref{fig:t-SNE} shows the performance of t-SNE algorithm with each features-family in several image datasets using Euclidean distance as similarity measure and 50 as perplexity value. Note how GSDP representations achieved the best performance on each features-family.
We observed that our GSDP object image representations are compactly grouped and have greater separation between categories than those t-SNE clustering built with high dimensionality features of CNN models (and its correspond PCA-reduced versions). 
Therefore, we can assume that our global semantic descriptor can construct object category representations distribution with the ability to maximize inter-class elements differences and minimize the intra-class differences. That is, with our approach, elements in each category must be as similar as possible, and elements in different groups must be as different as possible.

\begin{figure*}[h]
	\begin{center}
		\includegraphics[width=0.85\linewidth]{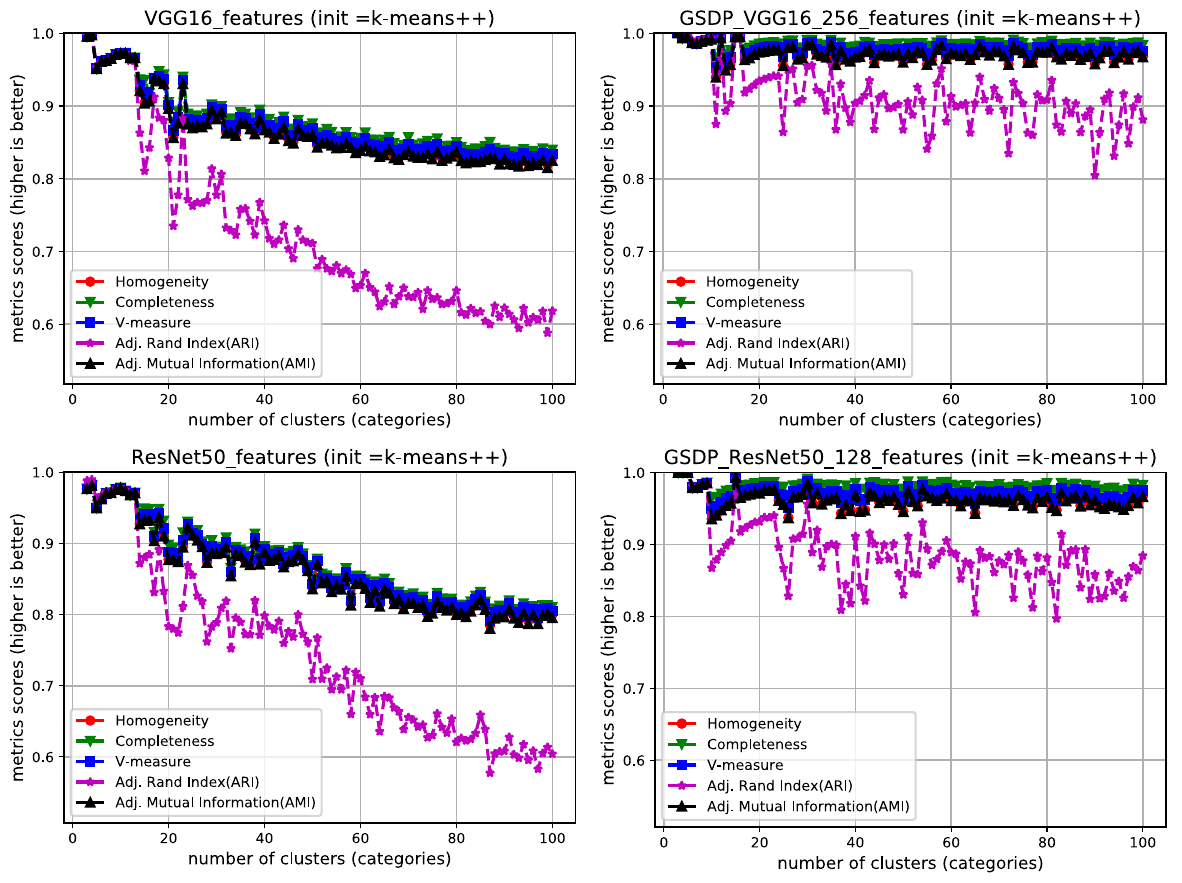}
		
	\end{center}
	\caption{History of K-Means metrics reached by each image feature representation in first 100 categories~(98 K-Means iterations) of ImageNet dataset. We compared the performance of VGG16 and ResNet50 features (left) \textit{versus} our GSDP descriptor signature (right) in clustering task.}
	\label{fig:Kmeans_metrics}
\end{figure*}


\subsubsection{Performance Evaluation}

\subsubsection*{Clustering}

\cite{yang2016joint} showed that when the image features representations achieve good metrics in image clustering task, it can generalize well when transferred to other tasks. Based on these assumptions, we evaluated our semantic GSDP encoding to verify its usefulness and suitability in image clustering task. We evaluated our GSDP descriptor~(version based in VGG16 and ResNet50 classification models) performance in clustering task with ImageNet dataset image. We compared our GSDP representation performance  against the following image global descriptors: GIST~\cite{oliva2001modeling}, LBP~\cite{ojala2002multiresolution}, HOG~\cite{dalal2005histograms}, Color64 \cite{li2007texture}, Color\_Hist \cite{song2004content}, Hu\_H \_CH \cite{haralick1973textural,hu1962visual,song2004content}, VGG16 features and  ResNet50 features (and its correspond PCA-reduced versions). 

We used K-Means algorithm for clustering $50,000$ images ($500\times$category) of first $100$ ImageNet dataset categories. The selection criteria of the K-Means algorithm is based on some similarities of the K-Means method with our image semantic representation approach. K-Means method minimizes the sum of squared errors between data points and their nearest cluster centers. This approach has similarities with our GSDP representation since GSDP signatures were constructed to organize features categories using as category organization center the abstract prototype signature.

We evaluated each image representations performance  in image clustering task  comparing its K-Means clustering metrics (Homogeneity, Completeness,
V-measure, Adjusted Rand Index, Adjusted Mutual Information).  For each global image representation, the experiment was conducted incrementally, starting with $3$ cluster (for $3$ categories) and incrementing a category for each K-Means algorithm iteration. At the end of each K-Means execution, the clustering metrics were saved. The idea behind our clustering experiment was to evaluate each image representation performance as the amount and diversity of objects images increased.


\begin{table}[]
	\caption{K-Means cluster metrics for each evaluated global image representation.
		Screenshot of K-Means measures for first $20$ ImageNet categories~($20$ clusters). We show Homogeneity~(H), Completeness~(C), V-measure~(V), Adjusted Rand Index~(ARI) and Adjusted Mutual Information~(AMI) clustering  measures. We show in bold the best performance.}
	\label{table:Kmeans_metrics}
	\resizebox{\columnwidth}{!}{%
		
		\begin{tabular}{|l|c|c|ccccc|}
			\hline
			\multicolumn{1}{|c|}{\multirow{2}{*}{\textbf{Descriptor}}} & \multirow{2}{*}{\textbf{Size}} & \multirow{2}{*}{\textbf{FPS}} &\multicolumn{5}{c|}{\textbf{Metrics Scores}}                                                                                   \\ \cline{4-8} 
			\multicolumn{1}{|c|}{} & & & \multicolumn{1}{c|}{H} & \multicolumn{1}{c|}{C} & \multicolumn{1}{c|}{V} & \multicolumn{1}{c|}{ARI} & AMI   \\ \hline \hline
			GIST      & 960& 0.82  & 0.05   & 0.05   & 0.05   & 0.01  & 0.05 \\                     
			LBP       & 512& 0.72  & 0.02   & 0.03   & 0.03   & 0.01  & 0.02 \\                     
			HOG       & 1960& 33 & 0.04   & 0.04   & 0.04   & 0.01  & 0.03   \\                     
			Color64   & 64& 8   & 0.12   & 0.12   & 0.12   & 0.04  & 0.11     \\                  
			Color\_Hist   & 512& 26  & 0.08   & 0.08   & 0.08   & 0.03  & 0.07 \\                   
			Hu\_H\_CH & 532& 6.9  & 0.04   & 0.04   & 0.04   & 0.01  & 0.02     \\    \hline              
			VGG16     & 4096& 15 & 0,87 &    0,88 &    0,88    & 0,78 & 0,87     \\                     
			VGG\_PCA\_256& 256& 12.5 &0,89    & 0,90    & 0,89    & 0,82    & 0,89 \\              
			GSDP\_VGG\_256& 256& 12.8  & \textbf{0,97} &    \textbf{0,99} &    \textbf{0,98} &    \textbf{0,93} &    \textbf{0,97} \\
			VGG\_PCA\_1024& 1024& 12.5 &0,89 &    0,89&    0,89&    0,81&    0,89 \\            
			GSDP\_VGG\_1024& 1024& 11.6  & 0,94&    0,98&    0,96&    0,84&    0,94 \\ \hline 
			ResNet50     & 2048& 10.6 &0,88&    0,90&    0,89&    0,78&    0,88    \\                     
			RNet\_PCA\_128& 128& 12.5 &0,88&    0,88&    0,88&    0,81&    0,88 \\              
			GSDP\_RNet\_128& 128& 9.6  &\textbf{ 0,97}&    \textbf{0,98}&    \textbf{0,98}&    \textbf{0,93}&    \textbf{0,97} \\
			RNet\_PCA\_512& 512& 12.5 &0,89&    0,90&    0,90&    0,82&    0,89 \\            
			GSDP\_RNet\_512& 512& 9  & 0,91&    0,97&    0,94&    0,73&    0,91 \\ \hline 
		\end{tabular}
	}
	
\end{table}

Table~\ref{table:Kmeans_metrics} shows a screenshot of K-Means clustering metrics achieved by each global image descriptor for the first $20$ ImageNet categories. Also, Table~\ref{table:Kmeans_metrics} shows features dimension and feature extraction velocity~(frame per second - FPS) for each image representation approach. All experiments were performed on a standard computer, without the use of GPUs to be fair with handcraft features approaches.

Note that our object image semantic representation achieved the best performance among the image representations evaluated with the ImageNet images sample. Also, be mindful of our GSDP descriptor representation keeps the same semantic information used by our CPM model for VGG16 and ResNet50 features interpretation (See Section \ref{subsection:signature_information} and Figure~\ref{fig:prototypical_organization}), but with more discriminatory image representation and even lower feature dimension. Experiments showed that lowest dimensional GSDP representations obtained the best cost-benefit performance.


Figure~\ref{fig:Kmeans_metrics} shows K-Means metrics history for VGG16 and ResNet50 features representation against the correspond GSDP signatures. We showed K-Means metrics behavior for each image representation when the number of clusters increases (until 100 categories) in each K-Means algorithm execution. Experiments showed that as object images variety increased, K-Means clustering metrics related to CNN-features deteriorated significantly, while K-Means clustering metrics achieved by our image semantic encoding remain above 0.9. Results showed that our semantic descriptor encoding significantly outperforms others image global encodings in terms of cluster metrics.

\subsubsection*{Classification}

To evaluate our image semantic encoding performance with supervised and unsupervised learning techniques, we also evaluated the performance of our GSDP representation in an image classification task.

%

\begin{figure*}
	\centering
	\subfigure[]
	{
		\includegraphics[width=0.83\linewidth]{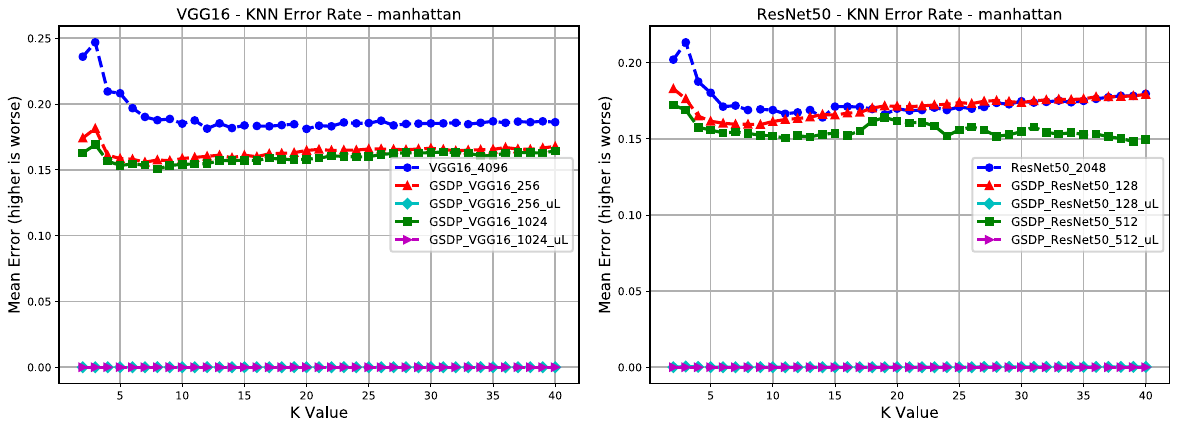}
		\label{fig:knn_manhatan} 
	}
	\\
	\subfigure[]
	{
		\includegraphics[width=0.83\linewidth]{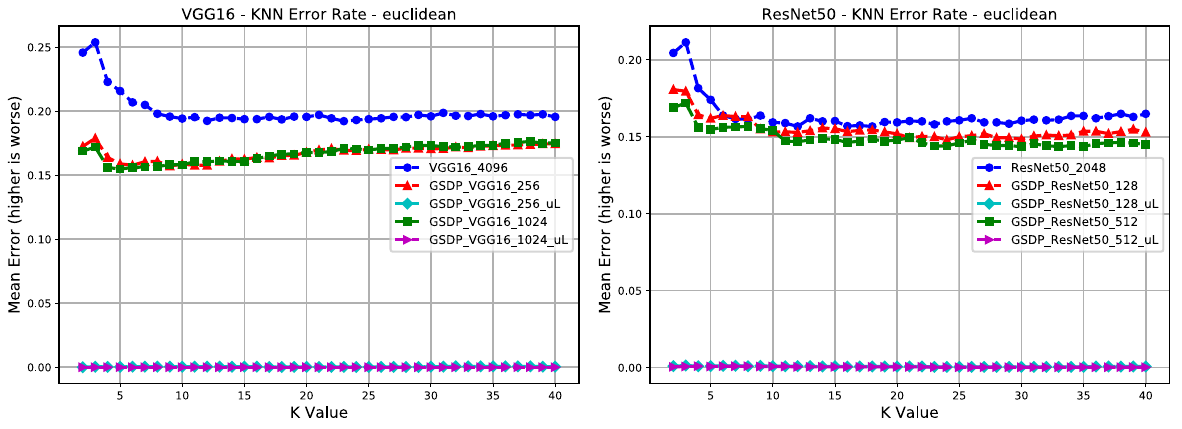}
		\label{fig:knn_euclidian} 
	}
	
	\caption
	{
		KNN error rate reached by each image representation in the first 100 categories of ImageNet dataset. We variated the K-value of KNN algorithm to compare the performance of VGG16 and ResNet50 features \textit{versus} our GSDP descriptor signature in image classification task using as feature similarity: \textit{a)}  Manhattan distance; \textit{b)} Euclidean distance. Each feature-length was placed in the corresponding caption.
	}

	\label{fig:knn_classification}
\end{figure*}

Our GSDP descriptor, by construction, builds objects image representations based on object category predictions made by CNN model used as background (See step depicted in  Figure~\ref{fig:methodology}b and Algorithm~\ref{alg:global_descriptor} line 4). Consequently, a prediction error of CNN-classification models generates that our descriptor constructs an object's image semantic representation using a wrong semantic prototype. This behavior is not problematic if we take into account that human beings will erroneously describe an object if it was previously wrong recognized.


In this experiment, we evaluated the performance of two GSDP semantic representations. Each GSDP semantic representation was constructed considering two different scenarios: \textit{i)} images GSDP-signatures are made based on the prediction of CNN model used in the background (normal behavior of our GSDP descriptor);
\textit{ii)} images GSDP-signatures are made based on the prediction of an ideal classification model~(100\% accuracy) (a hypothetical behavior of our GSDP descriptor). We used as a prediction of an ideal classification model, the image category label annotated in ImageNet dataset. We conducted the experiment to analyze the possible performance of our GSDP representation if the prototype selection error is zero (prediction error of CNN-model used in background). 


We performed our classification experiments using the KNN algorithm since, similar to t-SNE algorithm; elements are classified based on their local neighborhood. We analyzed the GSDP representation performance increasing the KNN algorithm parameter value (K neighbor) and using Euclidean and Manhattan distances as feature similarity measures.

Figure~\ref{fig:knn_classification} shows KNN algorithm performance using VGG16 and ResNet50 representations against corresponding GSDP signatures constructed in those two scenarios. In the experiment, we used the same ImageNet images sample used for clustering task evaluation. Also, we variated the K-value to show that our GSDP encoding significantly outperforms VGG16 and ResNet50 encodings in the KNN classification task. 

Experiments showed that our GSDP representation using the ResNet50 model reached a better performance than those constructed using the VGG16 model. Also, we observed that GSDP representations constructed using category labels~(notated with \_uL in Figure~\ref{fig:knn_classification}) are highly discriminative (mean error close to 0). Consequently, we can conclude that our semantic encoding of objects substantially improves its performance -- in classification task -- as the accuracy of the CNN-classification model used in background increases.

Our experiments showed that our image global semantic representation based on category prototypes could outperform other image global representations in some computer vision tasks. Also,  note that our GSDP encoding can describe objects images while encapsulates in its image signature the object semantic information (object semantic value and object typicality score).
Experiments showed that lowest dimensional GSDP representations (for each CNN model) were the ones that achieved the best size-performance trade-off.


\section{Limitations and Future works}


The proposed \textit{prototype-based description model} was constructed strictly to describe objects images, not to describe scenes images. Note that even when our semantic descriptor was evaluated in images that representing scenes, our GSDP descriptor can outperform other image global representations. The results achieved encourage us to evaluate the generalization ability of our object semantic representation in other computer vision tasks as image retrieval and scene understanding.

As future work, the interpretative criteria of human beings is necessary to construct images dataset with typicality annotations and conclude if our model can interpret objects images similar to human beings.

\section{Conclusion}

Motivated by how human beings represent and relate the meanings attributed to objects, this research was based on the Prototype Theory to propose semantic representations of object categories and object images. Specifically, in this paper we introduced and evaluated two models based on Prototype Theory foundations: \textit{i)} a Computational Prototype Model~(CPM) and \textit{ii)} a Prototype-based Description Model. 

We proposed the CPM model to represent the internal semantic structure of object categories. Experiments showed that our CPM model was able to encapsulate relevant features of objects category in our semantic prototype representation. Also, we showed that our semantic distance metric could simulate semantic relationships in terms of visual typicality, between category members. Our experiments showed that a relationship could be established between our semantic distance metric and object image visual representativeness. Expressly, our prototypical distance can be understood as the object image typicality score.
That is, our CPM model can capture the object's visual typicality and the central and peripheral meaning of objects categories. 

Based on the CPM model results, we proposed a prototype-based description model that uses the CPM model main components (\textit{semantic prototype} + \textit{semantic distance metric}) to construct a semantic representation of  object's image. Our prototype-based description model uses semantic prototypes of the CPM model to build a discriminatory signature that semantically describes object images highlighting its most distinctive features within the category.

Our novel Global Semantic Descriptor based on Prototypes~(GSDP)\footnote{All source code, prototypes datasets, and GSDP tool tutorials are publicly available in our lab's Github: https://github.com/verlab/gsdp.} introduces a new approach to the semantic description of object images. GSDP descriptor does not need to be trained, and it is easily adaptable to be used with any CNN-classification model. As shown in the experiments in the ImageNet dataset with VGG16 and ResNet50 models, our global semantic descriptor is discriminative, small dimensioned, and encodes the semantic information of category members. We further showed that our GSDP object representation preserves in its taxonomy the object's semantic meaning and the object typicality score.    

Our Prototype-based Description Model proposes a starting point to introduce the theoretical foundation related to \textit{the representation of semantic meaning} and \textit{the learning of visual concepts} of the Prototype Theory in the CNN semantic descriptors family.

\begin{acknowledgements}
This research was supported by funding from the Brazilian agencies: Coordena{\c c}{\~ a}o de Aperfei{\c c}oa- mento de Pessoal de N{\'i}vel Superior (CAPES), Conselho Nacional de Desenvolvimento Cient{\'i}fico e Tecnol{\' o}gico (CNPq), and Funda{\c c}{\~ a}o de Amparo a Pesquisa do Estado de Minas Gerais (FAPEMIG).
\end{acknowledgements}

%
%

\bibliographystyle{spbasic}      
\bibliography{reference}


\end{document}